\documentclass{article}

\usepackage[utf8]{inputenc}
\usepackage{newunicodechar}
\newunicodechar{─}{\rule[0.5ex]{1em}{0.55pt}}
 \usepackage[preprint]{neurips_2026}

\newif\ifshowcomments
\showcommentstrue

\usepackage[utf8]{inputenc} 
\usepackage[T1]{fontenc}    
\usepackage{hyperref}       
\usepackage{url}            
\usepackage{booktabs}       
\usepackage{amsfonts}       
\usepackage{nicefrac}       
\usepackage{microtype}      
\usepackage{xcolor}         
\usepackage{amsmath}
\usepackage{graphicx}
\usepackage{enumitem}
\usepackage[normalem]{ulem}

\usepackage{pgfplots}
\pgfplotsset{compat=1.18}
\usepgfplotslibrary{groupplots}
\usetikzlibrary{patterns}
\usepackage{xcolor}

\usepackage[capitalize, noabbrev]{cleveref}

\ifshowcomments
  \newcommand{\samuel}[1]{\textcolor{olive}{[SAMUEL: #1]}}
  \newcommand{\ohav}[1]{\textcolor{teal}{[Ohav: #1]}}
\else
  \newcommand{\samuel}[1]{}
  \newcommand{\ohav}[1]{}
\fi

\newcommand{\Simulation}{\textsc{Legion-SIM}}
\newcommand{\Benchmark}{\textsc{Legion-Bench}}

\usepackage[utf8]{inputenc} 
\usepackage[T1]{fontenc}    
\usepackage{hyperref}       
\usepackage{url}            
\usepackage{booktabs}       
\usepackage{amsfonts}       
\usepackage{nicefrac}       
\usepackage{microtype}      
\usepackage{xcolor}         

\title{LEGION: Benchmarking Language Agents on Sparse-Signal Investigation Under Adversarial Noise}

\author{%
  Samuel Joseph Amouyal \\
  Legion Security \\
  \texttt{samuel@legionsecurity.ai} \\
  \And
  Ofek Glick \\
  Legion Security \\
  \texttt{ofek@legionsecurity.ai} \\
  \AND
  Yaniv Menasherov \\
  Legion Security \\
  \texttt{yaniv@legionsecurity.ai} \\
  \And
  Eyal Fisher \\
  Legion Security \\
  \texttt{eyal@legionsecurity.ai}
  \And
  Ohav Barbi \\
  Legion Security \\
  \texttt{ohav@legionsecurity.ai} \\
}

\begin{document}

\maketitle

\begin{abstract}

Autonomous language-model agents are increasingly deployed on discovery-driven tasks where signal is sparse and most retrieved information is irrelevant. Current benchmarks fail to evaluate this regime due to dense signals,  cooperative environments, and ground truth typically being only a few tool calls away. We introduce \textsc{Legion}, a benchmark framework coupling a stochastic enterprise-telemetry simulator (\Simulation) with a tiered library of 74 expertly-crafted adversarial scenarios (\Benchmark) and an evaluation protocol that scores investigation efficiency alongside detection accuracy. Scenarios are injected into freshly simulated companies, allowing independent control over confounder density and event volume. We test five agent architectures on two simulated companies and find that while recall is high (0.82-0.88) precision lags behind (0.61--0.72). They fail to find all of the indicators of compromise of the attack or mistake genuine indicators with attack related ones, which is critical in the context of cybersecurity investigation. Failures are consistent across reasoning tiers and persist across confounder density and event volume. We release \textsc{Legion}, the scenario library, performance on common agent architectures, and pipeline for reproducibility and extension.
    
\end{abstract}
\section{Introduction}
\label{sec:intro}

Large language models ~\citep{gemini-1.5, gpt-4, qwen3.5} are rapidly transitioning from passive question-answering systems into autonomous agents that act in the world ~\citep{toolllm, gorillalm}. They write code ~\citep{codegen-survey}, navigate complex interfaces ~\citep{webagents-survey, example-webagent}, and orchestrate workflows across enterprise systems ~\citep{workflow-enterprise}. Agents are increasingly entrusted with decisions that carry real consequences: triaging medical records, flagging fraudulent transactions, and investigating security incidents ~\citep{ai-fraud, ai-medical-triage}. As the autonomy granted to these agents grows, a precise account of what they \emph{can} and \emph{cannot} do becomes not merely useful, but necessary.

LLM agents are typically evaluated on tool use and environment interaction ~\citep{humaneval, swe-bench, webarena, assistantbench, preventing}. These benchmarks share a structural property that quietly limits what they reveal: \emph{the environment is informationally cooperative}. A failing test pinpoints the bug ~\citep{swe-bench}; a web page surfaces the required elements ~\citep{webarena, webshop, assistantbench}; an API call returns ground truth on demand ~\citep{toolbench}. Relevant signal is dense, retrievable, and rarely disguised.

This is not true in real investigations. Most observations are distractors, evidence is obscured, and no single query is decisive. Success requires \textbf{abductive reasoning}: forming hypotheses from sparse evidence, designing queries to test them, and revising beliefs across interactions in which the overwhelming majority of retrieved information is noise. This an issue of \emph{sustained inference under uncertainty}. The challenge is twofold: an agent must find the relevant evidence while also not being misled by the irrelevant. Confidently flagging spurious elements is as much a failure as missing true ones. When an agent makes 50 tool calls and produces a report full of spurious indicators, the failure rarely localizes to a single step: early missteps propagate silently, defining the search space for every query that follows. We call this \emph{recursive discovery}, and existing benchmarks cannot isolate it because they fail to produce the conditions that trigger it: low signal-to-noise ratio, cross-source evidence fusion, and adversarial distractors that lead confident pattern-matching to wrong answers.

\begin{figure}[t]
\centering
\includegraphics[width=1.0\textwidth, height=5.5cm, trim={0cm 0cm 0cm 0cm}, clip]{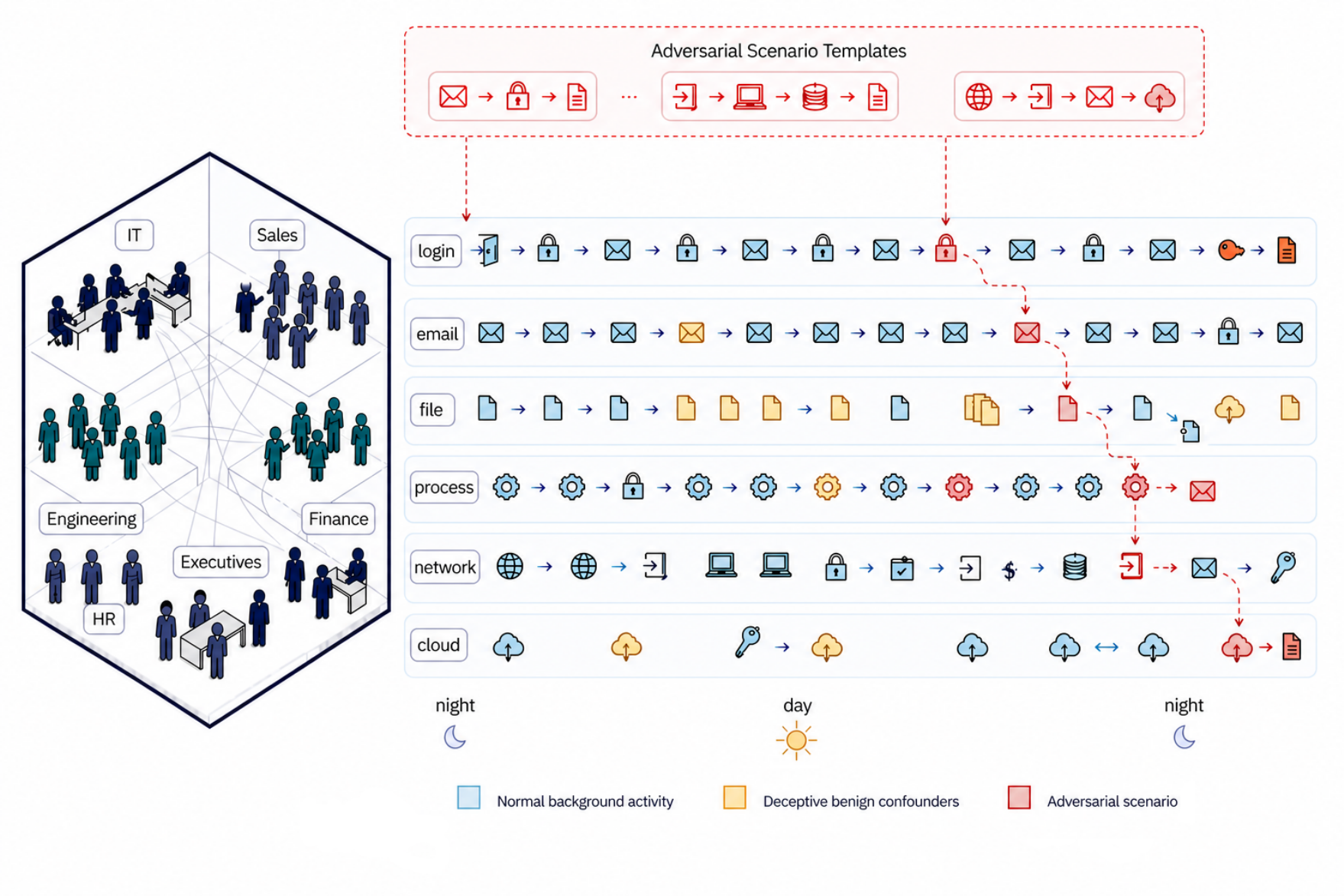}
\caption{The \Simulation \ pipeline. A stochastic enterprise environment generates event streams seeded with deceptive benign \emph{confounders}. Adversarial scenario templates are then causally injected into this stream yielding realistic investigations in which signal is structurally indistinguishable from noise without contextual reasoning.}
\setlength{\textfloatsep}{0.5pt}
\end{figure}


To this end, we introduce \textsc{Legion}, a framework designed to isolate and evaluate recursive discovery in enterprise security operations.
\textbf{\Simulation} is a stochastic environment generator that produces enterprise environments with thousands of employees and hundreds of thousands of heterogeneous events. Crucially, it injects \emph{deceptive benign activity}, confounders statistically indistinguishable from attack indicators without conditioning on each entity's role and baseline, so that naive pattern matching yields high false-positive rates and agents must reason over \emph{distributional context} rather than surface signals.
\textbf{\Benchmark} is a tiered evaluation suite of 74 attack scenario templates built on top of \Simulation, each specifying a causally ordered kill chain, ground-truth indicators of compromise, and expected investigative findings. Three tiers isolate distinct capabilities: \emph{Tier~1} tests retrieval of a known indicator from a noisy observation space; \emph{Tier~2} tests fusion of individually ambiguous signals across heterogeneous channels within an hour-scale window; \emph{Tier~3} tests sustained hypotheses across multiple days of dwell time, following lateral connections without contextual drift.

We evaluate five agent architectures (Direct, ReAct, See--Plan--Act, Reflexion, and Multi-Agent) along  \emph{recall} and \emph{precision}. Cybersecurity operations are an example of a domain in which both recall (retrieving all the attack event) and precision (distinguishing between genuine and malicious elements) are essential. While recall is high (0.82--0.88), precision lags behind (0.61--0.72). Additionally,  they are both far from perfect which in the context of cybersecurity investigation can be specially dangerous. Interestingly the expected cliff between single-event detection and multi-day campaign reconstruction is gentler than expected.


These results highlight a central challenge for autonomous agents in discovery-driven domains: improving precision while preserving and further increasing recall. Agents recover much of the attack evidence, yet still miss relevant events and include spurious findings. Adaptive querying and more careful assessment of retrieved evidence are promising directions for addressing these gaps. We ground this challenge in cybersecurity, but similar demands may arise in other domains where agents must discover and evaluate evidence in noisy environments under a finite query budget.
\section{Related Work}
\label{sec:related}

\paragraph{Agent benchmarks and the cooperative-environment assumption.}
The first generation of LLM-agent benchmarks established a now-standard template: a well-scoped task, an environment that exposes ground truth on request, and a verifiable terminal state. HumanEval~\cite{humaneval} and SWE-Bench~\cite{swe-bench} score code generation against executable tests; WebArena~\cite{webarena}, WebShop~\cite{webshop}, and Mind2Web~\cite{deng2023mind2web} score web navigation against deterministic DOM transitions; ToolBench~\cite{toolbench} and API-Bank~\cite{li2023api} score API orchestration against schema-checked returns. Relevant elements are a few hops away and distractors are not adversarial. \Simulation\ violates each of these assumptions by design.

\paragraph{Long-horizon and discovery-oriented evaluation.}
A second line of work targets long-horizon and partially observed environments. AssistantBench~\cite{assistantbench} and VeriWeb~\cite{veriweb} require persistent exploration across interdependent subtasks and report sharp degradation as horizon length grows. GAIA~\cite{mialon2023gaia} couples multi-hop reasoning and tool use; BrowseComp~\cite{wei2025browsecomp} stresses information-seeking under search-engine noise; AgentBench~\cite{liu2023agentbench} and OSWorld~\cite{xie2024osworld} broaden the surface to OS interaction. These benchmarks treat horizon length and tool diversity as first-class variables, and several study how errors compound across steps. In \Simulation, horizon length is not the central difficulty: every attack indicator has benign confounders statistically indistinguishable from it without conditioning on entity-specific baselines.

\paragraph{Abductive and hypothesis-driven reasoning.}
A separate thread evaluates the capacity to form, test, and revise hypotheses under incomplete information. AgentClinic~\cite{schmidgall2024agentclinic} casts diagnosis as sequential evidence gathering; RULEARN~\cite{rulearn} probes interactive rule induction; DiscoveryWorld~\cite{jansen2024discoveryworld} embeds scientific-method loops in a simulated laboratory. These works share our emphasis on \emph{abductive} rather than purely deductive reasoning, but their search spaces are typically small enough for a competent agent to enumerate hypotheses exhaustively. \Simulation\ contains on the order of $10^5$ heterogeneous events per scenario, and the cost of a wrong hypothesis is paid in query budget rather than a single penalized turn — which makes \emph{recursive discovery} measurable.

\paragraph{Security operations as an evaluation domain.}
Cybersecurity has emerged as a natural testbed for agentic capabilities, as SOC work involves low signal-to-noise ratios and adversarial deception. ExCyTIn-Bench~\cite{wu2025excytin} grounds agents in security telemetry and scores multi-hop evidence assembly; CTI-REALM~\cite{chakraborty2026cti} evaluates threat-intelligence synthesis across structured and unstructured sources; NYU CTF~\cite{shao2024nyu} and Cybench~\cite{zhang2024cybench} measure offensive capability via capture-the-flag tasks. Defensive benchmarks remain scarce, and most operate on real-world logs that are either small enough to admit memorization or so privacy-constrained that ground-truth attack labels are inferred. \Simulation\ is, to our knowledge, the first generator that synthesizes enterprise telemetry with \emph{causally injected} attack scenarios alongside benign confounders that are statistically engineered to defeat surface-level detection. This yields labeled ground truth at scale without sacrificing realism in the distribution of distractors.
\section{The \textsc{Legion} Testbed}
\label{sec:testbed}

Building a benchmark for discovery-driven reasoning requires placing the agent in conditions where abductive reasoning is not merely helpful but \emph{structurally necessary}.
We designed \textsc{Legion} around three principles that jointly create this condition: \newline
\emph{\textbf{Ecological adversariality}}: the environment is not cooperative or indifferent, but actively deceptive. Adversarial activity is deliberately crafted to resemble normal behavior, ensuring that naive pattern matching yields a high false-positive rate. \newline
\emph{\textbf{Diagnostic isolability}}: \emph{Legion} is stratified so that success at one tier paired with failure at the next yields a localized deficit signal, distinguishing whether an agent fails to find evidence, to filter noise, or to reconstruct a causal chain, rather than collapsing these into a single score. \newline
\emph{\textbf{Parametric control}}: every axis of difficulty, noise level, temporal span, evidence sources and adversarial sophistication is a tuneable parameter, enabling difficulty curves and scaling studies.

We instantiate these principles through two artifacts: \Simulation, a stochastic environment generator that produces enterprise environments with hundreds of thousands of heterogeneous events (\S\ref{sec:sim}), and \textsc{Legion-Bench}, a tiered evaluation suite that injects expertly-crafted adversarial scenarios of escalating complexity into that environment (\S\ref{sec:bench}).
Although we ground \textsc{Legion} in enterprise security operations, the capabilities it probes are domain-agnostic: any high-stakes setting that demands sustained reasoning under uncertainty will surface the same failure modes.

\subsection{\Simulation: A Stochastic Enterprise Environment}
\label{sec:sim}

Existing agent benchmarks present curated tasks within bounded action spaces. Real investigations require the agent to \emph{find} relevant leads in a massive, noisy observation space and folllow them. \Simulation \ generates such an environment: a stochastically evolving world producing hundreds of thousands of structured interconnected events, of which only a tiny fraction are task-relevant.
\paragraph{Company model.} Each company is specified by a YAML file defining an organizational graph: employees with role-conditioned parameters, departments, reporting hierarchies, devices, subnets, cloud infrastructure, mailboxes, and security groups. At runtime, we instantiate templates against this graph to produce ten event streams covering logins, emails, files, processes, applications, MFA, devices, prints, network flows, and administrative actions, plus three cloud streams for API calls, workload lifecycles, and threat-intelligence matches. All streams are materialized into a SQLite database tuned for the temporal, identity, and resource-access patterns of real SOC queries.
\paragraph{Employee-logs engine.} Employees act according to role-conditioned distributions with circadian patterns, weekly cycles, and episodic bursts. Activity volumes follow tunable Zipf-like distributions. Each employee carries a stable profile (fixed IPs, known devices, preferred MFA, a Zipf-weighted application distribution) plus a small personalized noise term, forcing the agent to reason about \emph{who} is acting, not just \emph{what} occurred. Process events are drawn from four buckets: file-correlated, application-correlated, role-specific, and background OS processes, with realistic parent--child trees, log-normal lifetimes, and circadian scheduled-task profiles. The engine also generates correlated multi-actor events (email threads, cross-department collaborations, coordinated administrative actions), with participants selected by organizational proximity.

\paragraph{Email and file generation.} Realistic content requires company-specific vocabulary, so we split generation into two phases. \emph{Offline}, an LLM is prompted with the company YAML to produce (i) a catalog of email scenarios 
with participant roles, subjects, bodies, and timing offsets, and (ii) a file catalog with realistic paths, sensitivity labels, owner roles, and content descriptions (see Appendix~\ref{app:gen_prompts} for the prompt). \emph{Online}, the stochastic engine samples subjects and bodies from the templates, recipients from a department-biased distribution (60/30/10 same/adjacent/random), and attachments on 40\% of messages from per-format size distributions. File events follow a session-based Markov chain with heavy-tailed sessions; action probabilities condition on the storage system (e.g.\ SharePoint biases toward reads), and shares spawn matching read tasks. Sensitivity labels propagate to the attack engine, which targets high-value files during exfiltration.

\paragraph{Network and cloud logs.} Network flows cover internal east--west and external north--south traffic across twelve protocol templates (HTTPS, DNS, and others), each carrying direction, byte volumes, authentication method, and geolocation. Cloud logs are generated per instance and container at role-dependent rates, capturing service discovery, IAM changes, storage access, and workload management. Together, these streams ensure attack evidence can be distributed across all telemetry.

\paragraph{Deceptive benign activity.} The engine deliberately produces benign events that resemble attack indicators, so naive pattern-matching yields high false-positive rates. We inject four \emph{confounders} targeting common detection heuristics: late-night sessions, travel logins, bulk file operations, and failed-login clusters. Each generates fully correlated multi-field events — a travel login, for instance, combines unfamiliar geography, an unknown-device flag, and a plausible foreign IP, indistinguishable from account compromise without reasoning about the employee's context. Each is injected at a configurable rate, capped at three per employee per window. Full company configuration details in Appendix~\ref{app:company_config}; schemas and example records appear in Appendix~\ref{app:log_schemas}.

\subsection{\Benchmark: A Tiered Reasoning Gauntlet}
\label{sec:bench}

\begin{table*}[t]
\centering
\scriptsize
\begin{tabular}{@{}lrrrrp{7.0cm}@{}}
\toprule
\textbf{Family} & \textbf{\#} & \textbf{T1} & \textbf{T2} & \textbf{T3} & \textbf{Representative templates} \\
\midrule
Social engineering / phishing       & 11 & 0 &  9 &  2 & Manipulate a user into malicious actions \\
Account takeover / identity         & 14 & 3 &  6 &  5 & Operate as a legitimate user via stolen credentials. \\
Endpoint / malware                  & 13 & 0 &  9 &  4 & Execute attacker code on a host to obtain persistent device control. \\
AD / hybrid identity                &  8 & 1 &  6 &  1 & Abuse Kerberos and directory primitives to escalate privilege. \\
Cloud / SaaS                        &  9 & 0 &  7 &  2 & Misuse cloud IAM, APIs, or admin actions. \\
Data theft / exfil / DLP            & 10 & 0 &  4 &  6 & Stage and egress sensitive data through monitored or covert channels. \\
Supply chain / DevOps               &  4 & 0 &  3 &  1 & Poison build pipelines, container images, or CI/CD credentials. \\
\midrule
Compound T3 campaigns               &  5 & 0 &  0 &  5 & Multi-stage chains spanning two or more families. \\
\midrule
\textbf{Total}                      & \textbf{74} & \textbf{4} & \textbf{44} & \textbf{26} & \\
\bottomrule \\
\end{tabular}
\caption{Threat-surface families in the \Simulation attack catalog and their distribution across reasoning tiers. T1: minutes-scale. T2: hour-scale. T3: multi-day campaign with identity chaining.}
\label{tab:attack_taxonomy}
\end{table*}
\Benchmark{} is built around the insight that a correct verdict requires a structured reconstruction over several data sources. It provides 74 scenario templates, crafted by a security expert to reflect realistic adversarial behavior. Each template is a parameterized, causally ordered chain of heterogeneous events - emails, logins, process executions, file operations, network flows, and cloud audit records — paired with a kill-chain narrative, ground-truth indicators of compromise (IoCs), and expected investigative findings. Templates are organised along two orthogonal axes.


\paragraph{Threat-surface families.} Templates are grouped into seven families that mirror how SOC teams triage incidents (Table~\ref{tab:attack_taxonomy}). The grouping is descriptive rather than a partition; its purpose is to report family-level performance and ensure coverage across major intrusion patterns rather than concentrating on a single style of attack (e.g.\ phishing).

\paragraph{Reasoning tiers.} Each template carries a tier label from \{T1,T2,T3\}, indicating the expected reasoning capability required to solve it (\S\ref{sec:bench}). Tiers are grounded in real-world dwell time: T1 scenarios are minutes-scale, single-action attacks with a direct indicator; T2 scenarios are hour-scale, single-session multi-phase chains whose individual events are ambiguous in isolation; T3 scenarios are multi-day campaigns with persistent state and identity chaining.

At test time, templates are instantiated against a concrete \Simulation\ company: victim identity, attacker infrastructure, and timing offsets are sampled per instance, yielding fresh ground truth for every run. Injection respects the infrastructure graph: an action targeting resource $m_j$ may only follow once the adversary holds credentials for $m_j$, and compromised entities continue their normal stochastic behavior between adversarial events, burying signal in noise. Attack events carry structured metadata (attack phase, victim identity, indicator flags) used exclusively by the evaluator and never exposed to the agent. See Appendix~\ref{app:attack_catalog} for the full templates descriptions.

\paragraph{Alert format.}
Each scenario instance has an \emph{alert} block as the agent's sole input. The information disclosed within the alert is minimal, removing the possibility of solving the task without investigation.

\begin{scriptsize}
\begin{verbatim}
### Description
Defender for Office 365 flagged suspicious inbound email activity targeting  
maverick.cox@meridianhcp.com  at 2026-04-03T16:49:38Z. The message matched heuristic 
detection rules. Investigate the full scope of the activity, identify indicators of 
compromise, and determine the impact.
### Initial Indicators
  - email: maverick.cox@meridianhcp.com
\end{verbatim}
\end{scriptsize}

\section{Experimental Setup and Evaluation}
\label{sec:experiments}
We evaluate a range of LLM agents on \textsc{Legion-Bench} to answer three main questions:
(1) Can agents reliably reconstruct a attack chains without surfacing spurious evidence?
(2) Does investigation quality degrade as complexity increases?
(3) Do the results vary across agent architectures?

\subsection{Tested Architectures}
\label{sec:agents}

Agents share a common infrastructure: they receive an alert with initial indicators, interact with the environment through the SOC tool interface (see Appendix ~\ref{app:soc_tools} for a full tool description) and produce a structured investigation report. They differ in the \emph{strategy} imposed during the investigation loop.

In the \textbf{Direct} architecture, the agent freely alternates between reasoning and tool calls until it produces a report. \textbf{ReAct}  \cite{yao2023react} adds structure by constraining the agent to a strict \emph{Thought--Action--Observation} loop, externalizing reasoning before every tool call. 
\textbf{SPA (See--Plan--Act)} \cite{assistantbench} enforces an alternation between a \emph{planning} phase and an \emph{execution} phase, with a \texttt{request\_replan} action allowing the agent to revise its plan mid-investigation. \textbf{Reflexion} \cite{shinn2023reflexion} performs two passes: after the first, a critic LLM prompted as a senior SOC analyst identifies gaps in the draft report, and the agent addresses them. Finally, \textbf{Multi-Agent} mirrors SOC team structure: a coordinator triages the alert and assigns tasks to domain specialists (endpoint, identity, email, data, threat intelligence, cloud), each with only the tools matching their domain, before synthesizing all findings into a unified report. See Appendix~\ref{app:prompts} for all our agents' prompts.

\paragraph{Golden Workflow.}
We compare our agents to a deterministic baseline that executes a fixed sequence of SOC tool calls derived from expert-authored playbooks (see Appendix~\ref{app:golden_workflow} for details). A \emph{single} LLM call synthesises the collected evidence into a report. Alerts are routed to one of four typed workflows (AccountTakeover, Phishing, Malware, DLP) based on the alert source field. The Golden Workflow serves as an interpretable reference point for assessing whether adaptive LLM reasoning adds measurable investigative value over deterministic automation.

\subsection{Evaluation Metrics}
\label{sec:metrics}

We evaluate agent performance along two complementary axes: \emph{the report} and \emph{the investigation}. This separation is essential, as a good report without finding all the attack events might miss the mark.

\paragraph{Reporting quality} We look at the Indicators of Compromise (IoC) surfaced in the report and measure \textbf{IoC Recall and Precision}. Each scenario specifies ground-truth IoCs grouped by category: IP addresses, domains, URLs, and email addresses. We evaluate IoC presence in the agent's report via string matching. Additionally, we measure \textbf{Findings Recall (FR).} Each scenario defines a set of \emph{expected findings}: specific conclusions a competent investigator should reach. We check whether the agent's \emph{Findings} section contains each one using an LLM-as-judge~\cite{llm-as-a-judge} and FR is the fraction of expected findings marked found. We validate our judge by manually inspecting 100 of the judgments and were in agreement with the judge in 98\% of the cases (see Appendix \ref{app:judge_prompt} for details).

\paragraph{Investigative coverage}
Each scenario carries attack events with full signatures. We can derive from there \textbf{Event Recall (ER)}: the fraction of attack events surfaced during the investigation. Unlike FR, ER scores the trajectory, not the write-up.



\section{Results}
\label{sec:results}

\begin{figure}[t!]
\centering
\includegraphics[width=\textwidth, height=5.5cm, trim={0.2cm 0.2cm 0.2cm 0.2cm}, clip]{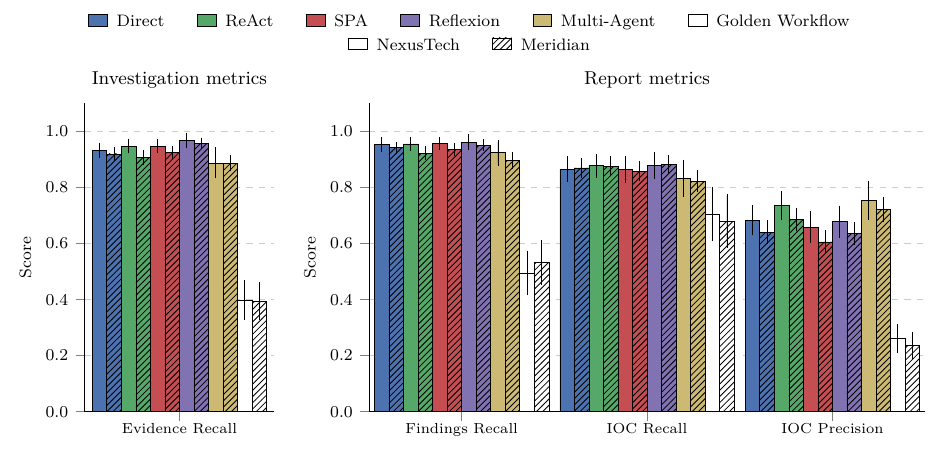}
\caption{Investigation metrics and report metrics for each agent on \textsc{NexusTech} (solid) and \textsc{Meridian} (hatched). Error bars are 95\% confidence intervals over scenarios.}
\label{fig:metrics-bars}
\end{figure}

\begin{table}[t]
\centering
\begin{minipage}[t]{0.52\linewidth}
\centering
\scriptsize
\begin{tabular}{@{}lcccc@{}}
\toprule
\textbf{Tier} & \textbf{ER} & \textbf{FR} & \textbf{IoC R.} & \textbf{IoC P.} \\
\midrule
T1 & 0.97 / 0.95 & 1.0 / 1.0 &  0.76 / 0.80 & 0.56 / 0.59 \\
T2 & 0.97 / 0.97 & 0.97 / 0.96 & 0.86 / 0.86 & 0.70 / 0.72 \\
T3 & 0.82 / 0.86 & 0.86 / 0.91 & 0.89 / 0.92 & 0.60 / 0.68 \\
\bottomrule \\
\end{tabular}
\caption{Per-tier performance averaged across all five agents. Format: \textsc{Meridian} / \textsc{NexusTech}.}
\label{tab:per_tier}
\end{minipage}
\hfill
\begin{minipage}[t]{0.46\linewidth}
\centering
\scriptsize
\begin{tabular}{@{}lcccc@{}}
\toprule
\textbf{Variant} & \textbf{ER} & \textbf{FR} & \textbf{IoC R.} & \textbf{IoC P.} \\
\midrule
Baseline                & 0.91 & 0.92 & 0.88 & 0.69 \\
$3{\times}$ confounders & 0.90 & 0.92 & 0.87 & 0.67 \\
$3{\times}$ events      & 0.89 & 0.92 & 0.86 & 0.65 \\
\bottomrule \\
\end{tabular}
\caption{Effect of scaling confounders and event volume on \textsc{Meridian}, ReAct only.}
\label{tab:scaling}
\end{minipage}
\end{table}

We evaluate all five agents, using GPT-5.5 (\cite{openai_gpt55_2026}) and a budget of up to 20 LLM calls, on the full \textsc{Legion-Bench} corpus, instantiated against \textsc{Meridian}, a healthcare company with 300+ employees and 600k events. We additionally run half the scenarios against \textsc{NexusTech}, a company with 2M+ events. Each template is instantiated five times per company, with results averaged.

\subsection{Neither recall nor precision are saturated}


\textbf{Frontier agents reach relative high recall and precision but never saturate.} 
Figure~\ref{fig:metrics-bars} compares report metrics (Findings Recall, IoC Precision and Recall) with investigation coverage (Event Recall) across all five architectures and both companies. Findings Recall (0.91--0.96), Event Recall (0.89--0.97), and IoC Recall (0.82--0.88) and Precision (0.61--0.72) are high but do not saturate---a critical limitation in cybersecurity, where identifying \emph{all} attack-related events and IoCs is essential.

\subsection{Per-tier behavior: the cliff is gentler than expected.}

Table~\ref{tab:per_tier} stratifies performance by reasoning tier. The expected recall cliff between cross-source correlation (T2) and multi-day campaigns (T3) is present but \textbf{shallower than expected}: 5--10 points of Findings Recall and 11--15 points of Event Recall separate T2 from T3, across agents. Interestingly, Tool-Call Accuracy rises with tier far less than event budget would suggest: T3 contains up to 3$\times$ more attack events than T1, yet TCA is essentially flat.

\subsection{Ablations.}
\label{sec:analysis_scaling}

\begin{table}[t]
\centering
\scriptsize
\resizebox{\linewidth}{!}{%
\begin{tabular}{@{}lcccccccccccc@{}}
\toprule
 & \multicolumn{4}{c}{\textbf{Direct}} & \multicolumn{4}{c}{\textbf{ReAct}} & \multicolumn{4}{c}{\textbf{Multi-Agent}} \\
\cmidrule(lr){2-5}\cmidrule(lr){6-9}\cmidrule(l){10-13}
\textbf{Model} & \textbf{ER} & \textbf{FR} & \textbf{IoC R} & \textbf{IoC P} & \textbf{ER} & \textbf{FR} & \textbf{IoC R} & \textbf{IoC P} & \textbf{ER} & \textbf{FR} & \textbf{IoC R} & \textbf{IoC P} \\
\midrule
GPT-5.6 Terra   & 0.83 & 0.89 & 0.79 & 0.75 & 0.82 & 0.90 & 0.79 & 0.72 & 0.75 & 0.85 & 1.00 & 0.67 \\
GPT-5.6 Luna    & 0.88 & 0.91 & 0.80 & 0.52 & 0.82 & 0.90 & 0.80 & 0.44 & 0.81 & 0.85 & 0.99 & 0.47 \\
Gemini Pro      & 0.78 & 0.82 & 0.75 & 0.71 & 0.76 & 0.84 & 0.82 & 0.65 & 0.64 & 0.37 & 0.94 & 0.85 \\
Claude Opus 4.7 & 0.75 & 0.89 & 0.82 & 0.56 & 0.75 & 0.87 & 0.85 & 0.61 & 0.71 & 0.91 & 0.99 & 0.40 \\
\bottomrule
\end{tabular}}
\caption{Agent performance on \textsc{Meridian} across backbone models. ER: Evidence Recall; FR: Findings Recall; IoC R / IoC P: IoC recall / precision. Runs that errored are excluded.}
\label{tab:different_models}
\end{table}

\begin{table}[t]
\centering
\scriptsize
\setlength{\tabcolsep}{3pt}
\begin{tabular}{@{}lcccccccc@{}}
\toprule
 & \multicolumn{4}{c}{\textbf{Direct}} & \multicolumn{4}{c}{\textbf{ReAct}} \\
\cmidrule(lr){2-5}\cmidrule(l){6-9}
\textbf{Toolset} & \textbf{ER} & \textbf{FR} & \textbf{IoC R} & \textbf{IoC P} & \textbf{ER} & \textbf{FR} & \textbf{IoC R} & \textbf{IoC P} \\
\midrule
Full toolset (baseline)       & 0.92 & 0.94 & 0.87 & 0.64 & 0.91 & 0.92 & 0.88 & 0.69 \\
No analysis tools     & 0.92 & 0.92 & 0.84 & 0.69 & 0.88 & 0.90 & 0.82 & 0.72 \\
Query-only & 0.85 & 0.86 & 0.81 & 0.66 & 0.84 & 0.89 & 0.83 & 0.70 \\
\bottomrule
\end{tabular}
\caption{Tool ablations on \textsc{Meridian} (GPT-5.5).}
\label{tab:tool_ablation}
\end{table}

\paragraph{Fixed playbooks} In both recall metrics and indicator F1 scores, LLM agents significantly outperform golden workflows. This shows that our benchmark is challenging, and that rule-following automations are not sufficient.

\paragraph{Additional confonders} The headline results in \S\ref{sec:results} use the baseline \Simulation\ parameters, raising a natural concern: would the apparent saturation on report-side metrics dissolve under heavier noise or larger event volumes? To test this, we re-instantiate \textsc{Meridian} along two perturbation axes with attack scenarios and agents held fixed: \texttt{confounders\_3x} triples the rate of benign red-herring activity, and \texttt{events\_3x} triples the per-employee event budget. Table ~\ref{tab:scaling} shows the results of each variant and compared to baseline ReAct runs. 
Overall, it does not significantly change the results, meaning that adding noise does not influence the model's performance.

\paragraph{Additional models} While we did not run the full setup (5 agents, with 5 victims for each scenario) due to the prohibitive cost, we did run Gemini-3.1-Pro, Opus-4.7, GPT-5.6-Luna and GPT-5.6-Terra on all the scenarios with 3 agents (React, Direct and SPA) with one victim for all the scenarios. In Table~\ref{tab:different_models} we can see that all these models do not saturate neither on recall nor on precision. Additionally, while recall is relatively similar across models, precision changes a lot, and strong models are essential to properly identify IoCs.

\paragraph{Tool ablations} To verify the impact of the tools provided on the investigation, we run two additional ablations. In the first, the model still received tools but we removed tools that conduct some detection logic (get risky users, detect suspicious processes, etc) that can be considered as being some oracle calls. Additionally, in threat intelligence, we reduce the information given to “Malicious/Benign” without noting which attacks the entity participated in. In the second, the agent is given only the ability to run SQL queries on the company database and the threat intelligence tool. In Table~\ref{tab:tool_ablation} we can see that the available tools do not influence the results too much, just the Evidence Recall a bit.
\section{Analysis}
\label{sec:analysis}

To further demonstrate how \Simulation{} and \Benchmark{} can be used, we conduct additional analyses of our results by looking at where the agents break. Two classes stand out, each diagnostic of a distinct reasoning weakness.

\paragraph{Slow drip exploitation}
Several T3 scenarios reveal a common failure mode: without a sharp anomaly, agents often miss the full scope of adversarial activity. This is most evident in BEC invoice diversion (38--75\% FR, 37--76\% ER), where legitimate-looking emails are spread over weeks, and also appears in malicious browser extensions (49\% ER, 68\% FR) and consent phishing with slow exfiltration (44\% ER, 61\% FR). In these cases, agents detect the initial suspicious event but fail to trace delayed downstream activity across channels.



\paragraph{Insider theft using legitimate credentials.}
Mass-download-on-resignation (FR 52--72\%, ER 50--66\%) and source-code-repo theft (FR 57--83\%, ER 56--83\%) remain challenging across architectures. Here, all actions use valid credentials and legitimate access, so detection depends on identifying behavioral anomalies; such as unusual volume, out-of-scope repositories, or off-hours activity; rather than any single malicious event. Agents do not reliably perform this broader statistical reasoning.

\section{Conclusion}
\label{sec:conclusion}
We introduced \textsc{Legion}, a benchmark framework pairing a stochastic enterprise simulator with a tiered attack scenario library and an evaluation protocol scoring both detection accuracy and investigation efficiency. Findings and Event Recall sit between 0.82–0.95, but precision remains low (0.61–0.72) across all agents, with many elements spuriously flagged. The binding constraint for current agents is not finding the right evidence, but doing so without surfacing noise, and current architectures, including those designed to plan and reflect, do not solve it. At this level of precision, autonomous deployment in real investigative settings would be difficult. While these problems were surfaced from a cybersecurity environment, we believe these conclusions would generalize to any task which requires abductive reasoning within a noisy environment. We release \textsc{Legion}, \Benchmark, the agents, and the pipeline so the community can extend and contest these findings.
\footnote{Code and data: \url{https://github.com/legion-security/legion-bench}}

\section*{Limitations}
\label{sec:limitations}

While we aimed to make our simulation as realistic as possible, we believe that short of taking actual logs from a company (which would face anonymization considerations) it is near impossible to provide a 1 to 1 simulation of an actual company. However, we still argue our simulation is useful enough to surface interesting conclusions. Additionally, due to the high cost of running these agents in this environment, we did not scale the number of companies nor the base LLMs tested. Finally, while the results plainly show agents fail at this task in ways that differ from humans, a larger human comparison is required. This type of comparison is lengthy, still underway and will provide additional insights into the performance of LLM based agent with respect to humans.
Alongside the defensive improvements possible through this environment, we must also consider the risk posed. In theory, attackers can use the environment to craft and simulate attacks on companies, measuring AI agent performance and optimizing their attacks to evade agentic investigations. While this is a consideration, we believe the option to pinpoint agentic faults and improve them outweighs this risk. 

\bibliographystyle{plainnat}
\bibliography{references}


\appendix

\section{Company Generation Prompts}
\label{app:gen_prompts}

The offline preparation pass (\S\ref{sec:sim}) calls an LLM (Claude Sonnet via LiteLLM) four
times per company to generate email scenarios and a file catalog.
All calls share a common preamble; specialised prompts extend it with task-specific directives.
Dynamic fields derived from the company YAML are shown in \texttt{<angle brackets>}.

\subsection*{Shared Preamble (injected into every generation call)}

\begin{small}
\begin{verbatim}
You are generating email scenario templates for a company simulation system.

## Company Profile
- Name: <company_name>
- Domain: <domain>
- Description: <description>
- Size: <N> employees
- Locations: <location_list>

## Departments and Roles
- <Dept A> (<pct>% of employees): Role1, Role2, ...
- <Dept B> (<pct>% of employees): Role1, Role2, ...
[...]

## Template Slot System
<slot documentation: {sender}, {recipient_1}, {date}, etc.>

## Key Requirements
1. Multi-participant conversations: 2-4 participants
2. Realistic timing: use offset_hours for response times
3. Natural conversation flow
4. Professional tone specific to this industry
5. Industry-specific language and terminology
6. Output: JSON array matching this schema (one object per scenario):

[
  {
    "id": "<company_id>_<unique_slug>",
    "name": "Human-readable scenario name",
    "description": "Brief description of the conversation",
    "num_participants": 2,                   // integer, 2-6
    "participant_roles": ["initiator", "responder"],  // drives slot aliases
    "same_department": true,                 // false for cross-dept scenarios
    "departments": ["<dept_name>"],          // lock to specific departments
    "requires_manager": false,              // true if one participant = manager
    "cross_department": false,              // true if participants from diff depts
    "requires_file": false,                 // true if a file attachment is needed
    "file_types": ["pdf", "docx"],          // allowed attachment extensions
    "num_external": 0,                      // external (non-company) participants
    "urgency": "normal",                    // "normal" | "high" | "critical"
    "weight": 1.0,                          // selection weight
    "tags": ["collaboration", "review"],
    "emails": [
      {
        "offset_hours": 0.0,          // float: hours after scenario start
        "sender_index": 0,            // index into participant list (0=initiator)
        "subject_template": "Review needed: {file_name}",
        "body_template": "Hi {participant_1_name}, ...",
        "recipient_indices": [],      // empty = all other participants
        "include_attachment": false,
        "reply_to_previous": false,   // prepends "Re: " to original subject
        "cc_indices": []              // participant indices to CC
      }
    ]
  }
]

Template slots available in subject_template and body_template:
  {participant_N_name}, {participant_N_email}  -- N-th participant (0-indexed)
  {<role>_name}, {<role>_email}               -- alias from participant_roles
  {file_name}, {file_description}, {file_type} -- attachment context
  {company_name}, {company_domain}, {department}
  {project_name}, {deadline}, {quarter}, {date_reference}
  {meeting_day}, {meeting_time}
\end{verbatim}
\end{small}

\subsection*{Email Scenarios — Per-Department}

\begin{small}
\begin{verbatim}
<shared preamble>

## Department: <dept_name>
Roles in this department: <role_list>

Generate 5 unique email scenarios specifically for the <dept_name>
department at <company_name>.

Each scenario should:
1. Feel authentic to someone in <dept_name> at a company like <company_name>
2. Involve 2-4 participants with realistic roles from this department
3. Have 3-5 emails showing natural conversation progression
4. Include at least one scenario with a file attachment
5. Vary in urgency (mostly normal, one or two high priority)
6. Use industry-specific terminology
7. Set "departments": ["<dept_name>"] for each scenario
8. Prefix each id with "<company_id>_"

Output only valid JSON, no additional text or explanation.
\end{verbatim}
\end{small}

Additional calls produce 8 cross-department scenarios (participants from 2--3 departments), 6
urgent scenarios (\texttt{urgency: "high"} or \texttt{"critical"}, short response offsets,
subject-line urgency markers), and 10 generic scenarios covering the remaining variety.

\subsection*{File Catalog — Per-Department}

\begin{small}
\begin{verbatim}
You are generating file metadata for a company simulation system.

## Company: <company_name>
<description>

## Department: <dept_name>
Roles: <dept> - Role1, <dept> - Role2, ...

## Task
Generate exactly <N> file entries for the <dept_name> department.

Requirements:
1. Realistic documents that people in <dept_name> at <company_name> would use
2. owner role format: "Department - Role" using exact names above
3. Paths start with /departments/<dept_slug>/
4. Realistic extensions: .pdf, .docx, .xlsx, .pptx, .csv, .md
5. Subdirectories for organisation (e.g. .../Reports/, .../Projects/)
6. Sensitivity mix: ~45% business, ~30% regular, ~15% confidential,
   ~10% unlabeled
7. Dates between 2022-01-01 and 2025-03-31
8. Content: 1-2 sentence description of what the file contains
9. Industry-specific and realistic for <company_name>

Output only valid JSON array, no additional text or explanation.
\end{verbatim}
\end{small}

A separate call generates shared company-wide files under \texttt{/shared/} (policies,
HR documents, brand materials, compliance guides) using the same schema, with
\texttt{owner role} drawn from any department.
\section{Company Configuration Schema}
\label{app:company_config}

Each simulated company is specified as a YAML configuration file.
The file drives the static generation pass and controls every structural aspect
of the company: headcount, organizational graph, group memberships, network
topology, application inventory, and behavioral baselines.
Below we show an abbreviated but representative excerpt from the
\textsc{Meridian Healthcare Partners} configuration, one of the two companies
used throughout our experiments.
Fields that accept long lists (first-name pools, role-manager chains,
full application catalogs) are truncated with \texttt{[\ldots]} for space.

\begin{small}
\begin{verbatim}
# ── Overall workforce parameters ─────────────────────────────
overall:
  employees: 300
  department_distribution:
    Clinical Operations: 0.30
    Nursing:             0.15
    Administration:      0.10
    IT:                  0.05
    Finance & Billing:   0.08
    HR:                  0.05
    Compliance & Risk:   0.04
    Pharmacy:            0.04
    Laboratory:          0.05
    Radiology:           0.03
    Patient Services:    0.03
    Executive:           0.03
    Facilities:          0.05
  role_distribution:            # per-department title weights
    Clinical Operations:
      Chief Medical Officer:  0.01
      Medical Director:       0.03
      Attending Physician:    0.20
      Resident:               0.15
      Nurse Practitioner:     0.12
      Physician Assistant:    0.10
      Medical Assistant:      0.20
      Clinical Coordinator:   0.05
      Surgeon:                0.04
      Hospitalist:            0.05
      Clinical Informaticist: 0.05
    Nursing:
      Chief Nursing Officer:        0.02
      Nursing Director:             0.05
      Nurse Manager:                0.10
      Charge Nurse:                 0.10
      Registered Nurse:             0.35
      Licensed Practical Nurse:     0.20
      Certified Nursing Assistant:  0.13
      Nurse Educator:               0.05
    # [... 11 additional departments omitted ...]
  role_managers:                # reporting-chain specification
    Clinical Operations:
      Chief Medical Officer:  CEO
      Medical Director:       Chief Medical Officer
      Attending Physician:    Medical Director
      Resident:               Attending Physician
      Nurse Practitioner:     Medical Director
      # [...]

# ── Company identity ─────────────────────────────────────────
company:
  id:     meridian_healthcare
  name:   Meridian Healthcare Partners
  domain: meridianhcp.com
  description: >
    Regional healthcare network operating hospitals, outpatient clinics,
    and specialty care centers in the northeastern United States. HIPAA-
    regulated, uses EHR systems extensively, and employs physicians,
    nurses, and allied health professionals alongside administrative
    and IT staff.
  locations:
    - US-BOS-HQ
    - US-HFD
    - US-PVD

# ── Groups (security boundaries + authorisation sets) ────────
groups:
  - id:   grp_physicians
    name: Physicians Group
    member_roles:
      - Clinical Operations - Attending Physician
      - Clinical Operations - Surgeon
      - Clinical Operations - Hospitalist
      - Clinical Operations - Medical Director
    authorisations:
      - Epic EHR - admin
      - Cerner - admin
      - Dragon Medical - operator
      - Imprivata - operator
      - Microsoft Teams - operator

  - id:   grp_it_admins
    name: IT Administrators
    member_roles:
      - IT - Chief Information Officer
      - IT - IT Director
      - IT - IT Manager
      - IT - Systems Administrator
    authorisations:
      - Okta - admin
      - Active Directory - admin
      - CrowdStrike - admin
      - Imprivata - admin
      - Microsoft 365 - admin
      - Epic EHR - admin

  - id:   grp_compliance_team
    name: Compliance and Risk Team
    member_roles:
      - Compliance & Risk - Chief Compliance Officer
      - Compliance & Risk - Compliance Director
      - Compliance & Risk - Privacy Officer
      - Compliance & Risk - Internal Auditor
    authorisations:
      - Compliancy Group - admin
      - Healthicity - admin
      - Epic EHR - reader
      - SharePoint - operator
      - DocuSign - operator
  # [... 13 additional groups omitted ...]

# ── Network topology ─────────────────────────────────────────
networks:
  - {id: net_bos_hq, cidr: 10.0.0.0/16, location: US-BOS-HQ,
     name: meridian-hq-boston}
  - {id: net_hfd,    cidr: 10.1.0.0/16, location: US-HFD,
     name: meridian-hartford}
  - {id: net_pvd,    cidr: 10.2.0.0/16, location: US-PVD,
     name: meridian-providence}

# ── Device fleet ─────────────────────────────────────────────
devices:
  count: 350
  network_distribution:
    net_bos_hq: 0.50
    net_hfd:    0.25
    net_pvd:    0.25

# ── Mailboxes ────────────────────────────────────────────────
mailboxes:
  distribution_lists:
    - {address: physicians@meridianhcp.com,
       members_from_group: grp_physicians,       delegate_count: 3}
    - {address: nursing@meridianhcp.com,
       members_from_department: Nursing,         delegate_count: 4}
    - {address: compliance@meridianhcp.com,
       members_from_group: grp_compliance_team,  delegate_count: 2}
    - {address: it@meridianhcp.com,
       members_from_department: IT,              delegate_count: 3}
    - {address: exec@meridianhcp.com,
       members_from_group: grp_c_suite,          delegate_count: 1}
    # [...]

# ── Application inventory ─────────────────────────────────────
applications:
  internal:
    - {name: Patient Portal,        type: patient_engagement, criticality: critical}
    - {name: Clinical Dashboard,    type: analytics,          criticality: critical}
    - {name: Lab Information System,type: laboratory,         criticality: high}
    - {name: Staff Intranet,        type: knowledge_base,     criticality: medium}
  external:   # SSO-integrated applications (28 total)
    - {name: Epic EHR,          type: ehr,                  sso: true}
    - {name: Cerner,            type: ehr,                  sso: true}
    - {name: MEDITECH,          type: ehr,                  sso: true}
    - {name: Microsoft 365,     type: productivity,         sso: true}
    - {name: Microsoft Teams,   type: communication,        sso: true}
    - {name: Okta,              type: idp,                  sso: true}
    - {name: CrowdStrike,       type: edr,                  sso: true}
    - {name: Imprivata,         type: clinical_sso,         sso: true}
    - {name: SharePoint,        type: file_storage,         sso: true}
    - {name: Workday,           type: hris,                 sso: true}
    - {name: Philips IntelliSpace, type: medical_imaging,   sso: true}
    - {name: Compliancy Group,  type: compliance,           sso: true}
    - {name: QGenda,            type: scheduling,           sso: true}
    - {name: Citrix,            type: virtual_desktop,      sso: true}
    # [...]

# ── Temporal model and weekly activity weights ────────────────
generation_config:
  location_timezones:
    US-BOS-HQ: -5.0
    US-HFD:    -5.0
    US-PVD:    -5.0
  temporal:
    window_days: 30
    circadian:
      concentration: 2.5      # von Mises concentration for intra-day bursts
    weekly_weights:            # Mon–Sun relative activity multipliers
      default:    [1.0, 1.0, 1.0, 1.0, 1.0, 0.15, 0.10]
      clinical:   [1.0, 1.0, 1.0, 1.0, 1.0, 0.70, 0.60]
      nursing:    [1.0, 1.0, 1.0, 1.0, 1.0, 0.80, 0.75]
      laboratory: [1.0, 1.0, 1.0, 1.0, 1.0, 0.50, 0.40]
      radiology:  [1.0, 1.0, 1.0, 1.0, 1.0, 0.40, 0.30]
      pharmacy:   [1.0, 1.0, 1.0, 1.0, 1.0, 0.50, 0.40]
      executive:  [1.0, 1.0, 1.0, 1.0, 1.0, 0.20, 0.08]
      it:         [1.0, 1.0, 1.0, 1.0, 1.0, 0.20, 0.10]

# ── Episodic bursts ───────────────────────────────────────────
  bursts:
    enabled: true
    types:
      - name: patient_surge
        probability: 0.03
        duration_days:        { mean: 5,  std: 2 }
        intensity_multiplier: { mean: 2.0, std: 0.4 }
        affected_event_types: [logins, app_events, email_events]
        department_bias: [Clinical Operations, Nursing]
      - name: regulatory_audit
        probability: 0.02
        duration_days:        { mean: 10, std: 3 }
        intensity_multiplier: { mean: 1.8, std: 0.3 }
        affected_event_types: [file_events, email_events, app_events]
        department_bias: [Compliance & Risk, Administration]
      - name: quarter_end
        probability: 0.04
        duration_days:        { mean: 5,  std: 1 }
        intensity_multiplier: { mean: 1.8, std: 0.3 }
        affected_event_types: [email_events, file_events, app_events]
        department_bias: [Finance & Billing, Administration]

# ── Confounder injection ──────────────────────────────────────
  confounders:
    enabled: true
    types:
      - name: late_night_work
        rate: 0.10
        eligible_roles: ["*Director*", "*VP*", "*Chief*",
                         "*Senior*", "*Manager*", "*Attending*", "*Surgeon*"]
        effect:
          duration_events: { mean: 3, std: 1 }

      - name: travel_login
        rate: 0.04
        eligible_roles: ["*VP*", "*Director*", "*Chief*",
                         "*Executive*", "*Sales*"]
        effect:
          duration_events:           { mean: 2, std: 1 }
          unknown_device_probability: 0.5

      - name: bulk_file_operation
        rate: 0.05
        eligible_roles: ["*Billing*", "*Compliance*", "*HR*",
                         "*Manager*", "*Director*", "*Health Information*"]
        effect:
          file_count_multiplier: { mean: 4.0, std: 1.5 }

      - name: failed_login_cluster
        rate: 0.06
        eligible_roles: ["*"]
        effect:
          failure_count:      { mean: 4, std: 2 }
          followed_by_success: true

# ── Per-role behavioral baselines (role_policies) ─────────────
# Each role carries 9 baseline channels. Four representative roles
# shown below — the full config defines baselines for all 60 titles.

role_policies:
  # ---------- Attending Physician ----------
  # Clinical shift: early morning, midday, and evening login windows.
  # Low VPN usage (mostly on-premises workstations). EHR and Dragon
  # Medical-heavy process profile. Normal admin action rate near zero.
  "Clinical Operations - Attending Physician":
    baselines:
      login_baseline:
        vpn_pct: { mean: 0.10, std: 0.04 }
        hours:   ["06:00-08:00", "11:00-13:00", "14:00-16:00", "18:00-20:00"]
        isps:    ["Comcast", "Charter", "AT&T"]
        avg_events_per_day: { mean: 3.0, std: 1.0 }
      email_baseline:
        avg_inbound_per_day:    { mean: 55,  std: 15 }
        avg_events_per_day:     { mean: 6.0, std: 2.0 }
        typical_attachments:    ["pdf", "docx", "xlsx", "jpg"]
      process_baseline:         ["chrome", "epic", "dragon", "outlook",
                                 "teams", "imprivata", "citrix"]
      process_event_baseline:
        avg_events_per_day:     { mean: 6.0, std: 2.0 }
      net_baseline:             ["epic.meridianhcp.com",
                                 "mail.meridianhcp.com",
                                 "uptodate.com",
                                 "pubmed.ncbi.nlm.nih.gov"]
      file_baseline:
        avg_events_per_day:     { mean: 9.0,  std: 3.0 }
      network_baseline:
        avg_events_per_day:     { mean: 9.0,  std: 3.0 }
      device_baseline:
        avg_events_per_day:     { mean: 2.0,  std: 0.7 }
      app_baseline:
        avg_events_per_day:     { mean: 10.0, std: 3.0 }
      admin_baseline:
        avg_events_per_day:     { mean: 0.03, std: 0.02 }

  # ---------- Registered Nurse ----------
  # Three-shift rotation (07:00, 15:00, 23:00 shift starts).
  # Almost exclusively on-site; VPN pct \approx 3%. Simpler process set
  # dominated by Epic, Vocera, and Imprivata (badge SSO).
  "Nursing - Registered Nurse":
    baselines:
      login_baseline:
        vpn_pct: { mean: 0.03, std: 0.02 }
        hours:   ["06:00-07:00", "14:00-15:00", "22:00-23:00"]
        isps:    ["Comcast", "Charter", "AT&T"]
        avg_events_per_day: { mean: 3.0, std: 1.0 }
      email_baseline:
        avg_inbound_per_day:    { mean: 25,  std: 8  }
        avg_events_per_day:     { mean: 6.0, std: 2.0 }
        typical_attachments:    ["pdf", "docx"]
      process_baseline:         ["epic", "vocera", "outlook",
                                 "imprivata", "chrome"]
      process_event_baseline:
        avg_events_per_day:     { mean: 6.0, std: 2.0 }
      net_baseline:             ["epic.meridianhcp.com",
                                 "mail.meridianhcp.com",
                                 "meridianhcp.com"]
      file_baseline:
        avg_events_per_day:     { mean: 9.0, std: 3.0 }
      network_baseline:
        avg_events_per_day:     { mean: 9.0, std: 3.0 }
      device_baseline:
        avg_events_per_day:     { mean: 2.0, std: 0.7 }
      app_baseline:
        avg_events_per_day:     { mean: 10.0, std: 3.0 }
      admin_baseline:
        avg_events_per_day:     { mean: 0.03, std: 0.02 }

  # ---------- IT Systems Administrator ----------
  # Elevated VPN usage (38%), broad hours, and high process event
  # rate reflecting scripting and admin tooling. Network event rate
  # 2× the clinical baseline. Meaningful admin action rate (0.5/day).
  "IT - Systems Administrator":
    baselines:
      login_baseline:
        vpn_pct: { mean: 0.38, std: 0.10 }
        hours:   ["07:00-10:00", "12:00-16:00", "18:00-22:00"]
        isps:    ["Comcast", "Charter", "AT&T", "Verizon Fios"]
        avg_events_per_day: { mean: 8.0, std: 3.0 }
      email_baseline:
        avg_inbound_per_day:    { mean: 50,  std: 12 }
        avg_events_per_day:     { mean: 7.0, std: 2.0 }
        typical_attachments:    ["pdf", "xlsx", "ps1", "sh", "log"]
      process_baseline:         ["chrome", "powershell", "terminal",
                                 "okta", "activedirectory",
                                 "imprivata", "citrix"]
      process_event_baseline:
        avg_events_per_day:     { mean: 14.0, std: 4.0 }
      net_baseline:             ["okta.meridianhcp.com",
                                 "ad.meridianhcp.com",
                                 "teams.microsoft.com",
                                 "meridianhcp.com"]
      file_baseline:
        avg_events_per_day:     { mean: 9.0,  std: 3.0 }
      network_baseline:
        avg_events_per_day:     { mean: 18.0, std: 6.0 }
      device_baseline:
        avg_events_per_day:     { mean: 4.0,  std: 1.2 }
      app_baseline:
        avg_events_per_day:     { mean: 15.0, std: 5.0 }
      admin_baseline:
        avg_events_per_day:     { mean: 0.5,  std: 0.2 }

  # ---------- Compliance Analyst ----------
  # Standard office hours, moderate VPN (15%). Compliance tooling
  # (Compliancy Group, Healthicity) and heavy file access — this role
  # is eligible for the bulk_file_operation confounder, so benign
  # clustered file reads can resemble an exfiltration pattern.
  "Compliance & Risk - Compliance Analyst":
    baselines:
      login_baseline:
        vpn_pct: { mean: 0.15, std: 0.06 }
        hours:   ["08:00-10:00", "12:00-14:00", "15:00-17:00"]
        isps:    ["Comcast", "Charter", "AT&T"]
        avg_events_per_day: { mean: 3.0, std: 1.0 }
      email_baseline:
        avg_inbound_per_day:    { mean: 40,  std: 10 }
        avg_events_per_day:     { mean: 6.0, std: 2.0 }
        typical_attachments:    ["pdf", "xlsx", "csv", "docx"]
      process_baseline:         ["chrome", "outlook", "teams",
                                 "compliancy-group", "healthicity", "excel"]
      process_event_baseline:
        avg_events_per_day:     { mean: 6.0, std: 2.0 }
      net_baseline:             ["mail.meridianhcp.com",
                                 "teams.microsoft.com",
                                 "sharepoint.com",
                                 "meridianhcp.com"]
      file_baseline:
        avg_events_per_day:     { mean: 9.0, std: 3.0 }
      network_baseline:
        avg_events_per_day:     { mean: 9.0, std: 3.0 }
      device_baseline:
        avg_events_per_day:     { mean: 2.0, std: 0.7 }
      app_baseline:
        avg_events_per_day:     { mean: 10.0, std: 3.0 }
      admin_baseline:
        avg_events_per_day:     { mean: 0.03, std: 0.02 }
  # [... 56 additional role policies omitted ...]
\end{verbatim}
\end{small}

\noindent The full configuration for \textsc{Meridian Healthcare Partners} spans approximately 2\,800 lines covering 13 departments, 60 role titles, 16 groups, 3 office networks, 28 external applications, and complete \texttt{role\_policies} baselines for every role title.
The four roles shown above illustrate the principal axes of behavioral variation: VPN usage (3\%--40\%), login hour distributions (three-shift clinical vs.\ standard office hours), process event rates (6--14 events/day), network activity rates (9--18 events/day), and admin action rates (0.03--0.5 events/day).
These per-role baselines, combined with a small per-employee noise term sampled at generation time, are what force the agent to reason about \emph{who} is acting rather than merely \emph{what} action occurred.

\section{Event Log Schemas and Examples}
\label{app:log_schemas}

The simulator produces ten distinct event types, each stored in its own table
in the SQLite twin database.
Below we show one or two representative rows per type drawn from the
\textsc{NexusTech Global} baseline fixture. Attack-injected events carry an additional \texttt{\_attack\_metadata} field that is stripped before SQL insertion and used only for ground-truth evaluation.

\subsection*{Authentication Events (\texttt{logins})}
\begin{small}
\begin{verbatim}
{
  "ts":           "2026-03-27T00:25:53Z",
  "ip":           "39.161.90.94",
  "method":       "SSO+MFA",          // SSO | Password | Password+MFA
                                      // | SessionToken | MFA-only
  "success":      true,
  "app":          "DocuSign",
  "geo":          "VPN-US",
  "user_agent":   "Mozilla/5.0 (Windows NT 10.0; Win64; x64)
                   AppleWebKit/537.36 Chrome/124.0.0.0 Safari/537.36",
  "device_id":    "dev_00000",
  "known_device": true
}
// Attacker session replay (injected event):
{
  "ts":           "2026-03-27T01:42:10Z",
  "ip":           "185.234.72.143",
  "method":       "SessionToken",
  "success":      true,
  "app":          "Microsoft 365",
  "geo":          "Eastern Europe",
  "device_id":    "unknown",
  "known_device": false,
  "user_agent":   "python-requests/2.28.0"
}
\end{verbatim}
\end{small}

\subsection*{Email Events (\texttt{email\_events})}
\begin{small}
\begin{verbatim}
{
  "ts":               "2026-04-13T11:38:00Z",
  "direction":        "sent",          // sent | received
  "external":         false,
  "from":             "greta.long@meridianhcp.com",
  "to":               ["avery.ramos@meridianhcp.com"],
  "cc":               [],
  "bcc":              [],
  "subject":          "Sharing: executive_team_meeting_october2024.docx",
  "content":          "Hi Avery, sharing the October exec meeting minutes ...",
  "attachment_types": ["docx"],
  "attachments": [{
    "filename":  "executive_team_meeting_october2024.docx",
    "url":       "https://mail.meridianhcp.com/attachments/937641fdde84",
    "file_path": "/departments/executive/Meeting_Minutes/
                  executive_team_meeting_october2024.docx",
    "size_bytes": 162956
  }]
}
// Phishing lure (injected):
{
  "ts":           "2026-04-13T09:15:22Z",
  "direction":    "received",
  "external":     true,
  "from":         "security@login-secure-m365.com",
  "subject":      "Action Required: Verify Your Identity to Continue",
  "content":      "We detected suspicious activity on your account ...",
  "contains_url": "https://login-secure-m365.com/view/42891"
}
\end{verbatim}
\end{small}

\subsection*{File Events (\texttt{file\_events})}
\begin{small}
\begin{verbatim}
{
  "ts":         "2026-03-27T01:11:11Z",
  "action":     "read",            // read | modify | upload | delete | copy
  "path":       "\\\\docrepo\\Clinical Operations\\Reports\\
                  or_utilization_analysis_aug2024.pdf",
  "system":     "document_repository",
  "label":      "Business",        // DLP sensitivity label
  "owner_role": "Clinical Operations - Surgeon",
  "category":   "business",
  "description":"OR utilization report: surgical volume trends and
                 capacity optimization."
}
{
  "ts":         "2026-03-27T12:11:35Z",
  "action":     "modify",
  "path":       "https://meridianhealthcarepartners.sharepoint.com/
                 sites/Clinical Operations/Shared Documents/Reports/
                 quality_metrics_dashboard_q4_2024.xlsx",
  "system":     "microsoft_sharepoint",
  "label":      "Business",
  "owner_role": "Clinical Operations - Chief Medical Officer",
  "category":   "business",
  "description":"Quality metrics dashboard: patient safety indicators,
                 readmission rates, and clinical outcomes Q4 2024."
}
\end{verbatim}
\end{small}

\subsection*{Process Events (\texttt{process\_events})}
\begin{small}
\begin{verbatim}
{
  "ts":              "2026-03-29T14:08:52Z",
  "process_name":    "chrome.exe",
  "action":          "start",      // start | stop | crash | inject
  "parent_process":  "explorer.exe",
  "path":            "C:\\Program Files\\Google\\Chrome\\Application\\
                      chrome.exe",
  "pid":             24041,
  "integrity_level": "medium",
  "command_line":    "\"C:\\Program Files\\Google\\Chrome\\Application\\
                      chrome.exe\"",
  "hash":            "098f63fb41e1d75261742ba2044fe043617d3fcb1fd157b9...",
  "user_context":    "MERIDIAN\\greta.long",
  "device_id":       "dev_00300"
}
// Malicious process (injected):
{
  "ts":              "2026-03-29T14:42:17Z",
  "process_name":    "IT_Support_Tool.exe",
  "action":          "start",
  "parent_process":  "OUTLOOK.EXE",
  "path":            "C:\\Users\\Public\\Downloads\\IT_Support_Tool.exe",
  "integrity_level": "high",
  "command_line":    "IT_Support_Tool.exe --silent --install",
  "hash":            "e7b1c9d2a4f83b...",
  "user_context":    "MERIDIAN\\hans.ortiz",
  "device_id":       "dev_00142"
}
\end{verbatim}
\end{small}

\subsection*{Application Events (\texttt{app\_events})}
\begin{small}
\begin{verbatim}
{
  "ts":            "2026-03-27T12:09:57Z",
  "app":           "SharePoint",
  "action":        "export_data",  // access | login | export_data
                                   // | create_inbox_rule | search_mailbox
                                   // | open_url | navigate | ...
  "usage":         "read",
  "success":       true,
  "resource_type": "export",
  "resource_name": "Q4 Quality Metrics Report",
  "description":   "Exported SharePoint record to local download"
}
// Attacker action (injected):
{
  "ts":            "2026-03-27T23:14:03Z",
  "app":           "Outlook Web Access",
  "action":        "create_inbox_rule",
  "resource_name": "Auto-Delete Security Alerts",
  "description":   "Inbox rule to auto-delete emails containing
                    'security alert' or 'unusual sign-in'"
}
\end{verbatim}
\end{small}

\subsection*{MFA Events (\texttt{mfa\_events})}
\begin{small}
\begin{verbatim}
{
  "ts":          "2026-03-30T17:23:59Z",
  "action":      "push_sent",   // push_sent | push_approved | push_denied
                                // | totp_success | totp_failure
                                // | device_registered | device_removed
  "method":      "push",
  "device_name": "Okta Verify",
  "app":         "Okta",
  "ip":          "72.33.80.49"
}
{
  "ts":          "2026-03-30T17:24:06Z",
  "action":      "push_approved",
  "method":      "push",
  "device_name": "Okta Verify",
  "app":         "Okta",
  "ip":          "72.33.80.49",
  "success":     true
}
\end{verbatim}
\end{small}

\subsection*{Device Events (\texttt{device\_events})}
\begin{small}
\begin{verbatim}
{
  "ts":          "2026-03-27T00:09:14Z",
  "action":      "bluetooth_connect",  // bluetooth_connect | usb_insert
                                       // | usb_remove | device_enrolled
                                       // | device_wiped | screen_lock
  "device_type": "bluetooth_speaker",
  "vendor":      "JBL",
  "product":     "Clip 4",
  "device_id":   "dev_00000"
}
// USB exfiltration (injected):
{
  "ts":          "2026-03-27T18:33:41Z",
  "action":      "usb_insert",
  "device_type": "usb_storage",
  "vendor":      "SanDisk",
  "product":     "Ultra USB 3.0",
  "serial":      "48291037"
}
\end{verbatim}
\end{small}

\subsection*{Administrative Actions (\texttt{admin\_actions})}
\begin{small}
\begin{verbatim}
{
  "ts":          "2026-04-04T01:22:07Z",
  "system":      "okta",           // okta | azure_ad | active_directory
                                   // | aws_iam | endpoint_mgmt | ...
  "action":      "add_to_group",   // add_to_group | remove_from_group
                                   // | enable_mfa | register_mfa_device
                                   // | assign_role | create_service_principal
  "approved_by": "self",
  "details": {
    "group_id":   "grp_radiology_team",
    "group_name": "Radiology Team"
  }
}
// Persistence action (injected):
{
  "ts":            "2026-04-04T02:11:55Z",
  "system":        "okta",
  "action":        "register_mfa_device",
  "approved_by":   "self",
  "resource_name": "Microsoft Authenticator",
  "description":   "New Authenticator app registered from attacker session"
}
\end{verbatim}
\end{small}

\subsection*{Network Events (\texttt{network\_events})}
\begin{small}
\begin{verbatim}
{
  "ts":               "2026-03-27T00:03:18Z",
  "source_ip":        "10.10.0.244",
  "destination_ip":   "9.220.215.99",
  "destination_port": 80,
  "protocol":         "TCP",
  "action":           "allowed",   // allowed | blocked | detected
  "direction":        "outbound",
  "user":             "hao.gomez",
  "auth":             "none",
  "bytes_sent":       9903,
  "bytes_received":   9253,
  "description":      "HTTP web browsing"
}
// C2 beacon (injected):
{
  "ts":               "2026-03-29T22:04:51Z",
  "source_ip":        "185.234.72.143",
  "destination_ip":   "52.96.0.40",
  "destination_port": 443,
  "user":             "hans.ortiz",
  "bytes_sent":       524288,
  "bytes_received":   2097152,
  "description":      "Sustained HTTPS session to M365 from foreign IP —
                       mailbox and portal access"
}
\end{verbatim}
\end{small}

\subsection*{Print Events (\texttt{print\_events})}
\begin{small}
\begin{verbatim}
{
  "ts":            "2026-04-02T09:14:33Z",
  "document_name": "Patient_Discharge_Summary_Batch_April.pdf",
  "printer":       "HP-LaserJet-NursingStation-3W",
  "pages":         24,
  "copies":        1,
  "label":         "Confidential",
  "app":           "Epic EHR"
}
\end{verbatim}
\end{small}

\section{Attack Catalog}
\label{app:attack_catalog}

\textsc{Legion-Bench} ships 74 adversarial scenario templates spanning seven
threat-surface families.
Each template is parameterised (attacker domains, IPs, lure subjects, C2
infrastructure) and injected independently per scenario instance, so no two
runs share the same concrete IOCs.
The full catalog is listed below, grouped by reasoning tier.

\subsection*{Tier~1 — Single-indicator, pattern-matching difficulty (4 scenarios)}

\begin{description}[leftmargin=0pt, labelindent=0pt, itemsep=4pt]

\item[\textit{New MFA Device Registration for Persistence.}]
After compromise, attacker enrolls a new MFA method to maintain access even after a password reset and browses the victim's mailbox for sensitive content.

\item[\textit{Silent Mailbox Forwarding Rule.}]
Attacker logs in from a foreign IP, creates a silent forwarding rule to exfiltrate all incoming email to an external address.

\item[\textit{Inbox Hiding Rules for Alert Suppression.}]
Attacker creates inbox rules to suppress security-alert and finance-related emails and hide traces of the compromise.

\item[\textit{LSASS Memory Dump for Credential Extraction.}]
Attacker dumps LSASS memory to extract NTLM hashes and Kerberos tickets for subsequent lateral movement.

\end{description}

\subsection*{Tier~2 — Cross-source correlation, multi-step within a single session (44 scenarios)}

\begin{description}[leftmargin=0pt, labelindent=0pt, itemsep=4pt]

\item[\textit{AiTM Phishing --- Session Token Theft via Reverse Proxy.}]
Reverse-proxy between victim and real login captures session cookie; attacker replays stolen session, registers MFA device, creates inbox rules, and sets up mail forwarding.

\item[\textit{Device Code Phishing --- OAuth Token Hijack.}]
Attacker initiates Device Code OAuth flow and sends code to victim; victim unwittingly grants persistent token access; attacker browses portal, dumps mailbox, and registers MFA device.

\item[\textit{OAuth Consent Phishing --- Malicious App Authorization.}]
Victim authorises a malicious OAuth app with delegated Graph API permissions; app reads mailbox, accesses OneDrive, and adds a client secret for persistence.

\item[\textit{Email Thread Hijacking --- Trusted Reply Lure.}]
Attacker hijacks a legitimate email thread from a compromised peer's mailbox to lure a second victim into credential submission.

\item[\textit{Cloud Document Share Notification Lure.}]
Fake OneDrive sharing notification redirects victim to credential phishing page; attacker logs in from foreign IP and exfiltrates SharePoint files.

\item[\textit{Password Spraying --- Low-and-Slow Credential Attack.}]
Common passwords are tried across many accounts to avoid lockouts; successful login is followed by mailbox browsing and contact export.

\item[\textit{Privileged Role Assignment to Compromised Account.}]
Attacker with compromised admin credentials assigns Global Administrator role and creates a service principal for persistent elevated access.

\item[\textit{ISO/LNK Attachment Payload Chain.}]
Malware delivered via ISO disk image and LNK shortcut; payload beacons to C2, performs host discovery, and stages data for exfiltration.

\item[\textit{DLL Sideloading via Legitimate Application Bundle.}]
Phishing email delivers ZIP with a legitimate signed app and malicious DLL; sideloaded DLL establishes C2 and sets persistence via scheduled task.

\item[\textit{Legitimate Remote Access Tool Installation.}]
Attacker installs AnyDesk or similar RAT for persistent interactive access that blends with normal IT usage, then harvests credentials and exfiltrates data.

\item[\textit{Browser and Windows Credential Store Stealer.}]
Infostealer fingerprints the host, harvests credentials from browsers and DPAPI, compresses loot into an encrypted archive, and exfiltrates to C2.

\item[\textit{EDR / Security Tooling Disabling and Tampering.}]
Attacker enumerates security tools, disables EDR and Windows Defender, clears event logs, deploys payload, and maintains C2.

\item[\textit{Protocol Tunneling for Covert C2 Channel.}]
Encrypted reverse tunnel (chisel) over HTTPS blends C2 with web traffic; attacker performs host discovery, credential harvesting, and exfiltration through the tunnel.

\item[\textit{Kerberoasting --- Service Ticket Offline Cracking.}]
Attacker requests Kerberos TGS tickets for SPN-bearing accounts and cracks them offline to recover service account credentials.

\item[\textit{Pass-the-Hash NTLM Lateral Movement.}]
Stolen NTLM hash is injected to authenticate across multiple hosts via SMB without cleartext password; data collected on final target.

\item[\textit{DCSync --- Directory Replication Credential Extraction.}]
Attacker uses DRSUAPI replication rights to extract domain admin, krbtgt, and service account hashes.

\item[\textit{Golden Ticket --- Forged Kerberos TGT.}]
Using stolen KRBTGT hash, attacker forges TGTs to impersonate any domain identity and access high-value services.

\item[\textit{Local Admin Credential Reuse --- Mass Lateral Movement.}]
Shared local admin hash reused across many machines via SMB pass-the-hash with discovery commands at each hop.

\item[\textit{Cloud Service Discovery / Enumeration.}]
After gaining cloud access, attacker performs a broad resource sweep across subscriptions, storage, IAM, databases, and containers.

\item[\textit{Cloud Admin Run-Command Abuse.}]
Attacker uses AWS Systems Manager RunCommand to execute recon and exfiltration on production VMs without SSH/RDP access.

\item[\textit{Object Storage Mass Download.}]
Attacker enumerates S3 buckets, checks ACLs, and downloads large volumes of data using built-in sync tools.

\item[\textit{Cloud Secret Vault / Key Manager Theft.}]
Attacker enumerates key vaults and reads high-value secrets (database passwords, API keys, certificates).

\item[\textit{IAM Role / Policy Tampering for Privilege Escalation.}]
Attacker modifies IAM policy to add wildcard permissions and attaches them to their own identity for unrestricted access.

\item[\textit{Cloud Logging / Audit Trail Suppression.}]
Attacker disables CloudTrail, log sinks, and monitoring alerts before exfiltrating data under reduced visibility.

\item[\textit{SaaS Platform Bulk Data Export.}]
Attacker with compromised admin credentials initiates multiple export jobs (mailbox, SharePoint, Teams) and transfers them externally.

\item[\textit{Collaboration Platform Lure / Internal Phishing.}]
Compromised account scouts the org via Teams/Slack and sends targeted lures to harvest a second set of credentials.

\item[\textit{Source Code Repository Theft.}]
Developer credentials used to enumerate and clone multiple private repositories, scan for embedded secrets, and exfiltrate build artifacts.

\item[\textit{Poisoned CI/CD Pipeline via Malicious PR.}]
Attacker PR injects malicious workflow that dumps CI secrets, tampers with artifact integrity, and publishes a backdoored artifact.

\item[\textit{CI/CD Token Theft for Multi-System Access.}]
Malicious CI job dumps environment variables with cloud credentials, accesses Secrets Manager, and exfiltrates to attacker-controlled endpoint.

\item[\textit{Malicious Container Image Swap.}]
Attacker pushes a malicious image overwriting a legitimate container tag; Kubernetes pulls it; attacker executes commands and steals service account tokens.

\item[\textit{DocuSign Credential Harvesting.}]
Phishing email impersonating DocuSign redirects victim to fake login; attacker replays credentials from foreign IP, browses mailbox, and exfiltrates SharePoint documents.

\item[\textit{Microsoft 365 Credential Harvesting.}]
Attacker impersonates M365 security alert to steal credentials; uses them to browse portal, read mailbox, and download OneDrive files.

\item[\textit{IT Support Impersonation Malware Delivery.}]
Attacker impersonates IT support to convince victim to run malware disguised as a security update; malware establishes C2 and exfiltrates data.

\item[\textit{Voicemail Notification MFA Capture.}]
Fake voicemail HTML attachment presents an MFA capture page; stolen session replayed from foreign IP with subsequent inbox rule creation.

\item[\textit{Vendor Invoice QR Code Redirect.}]
Fake vendor invoice PDF contains QR code leading to a credential harvesting page; stolen credentials used to access mailbox.

\item[\textit{Credential Stuffing with Impossible Travel.}]
Breach-leaked credentials tested in rounds with cooldown; successful Evilginx session replay from geographically impossible location.

\item[\textit{MFA Fatigue Push Bombing Attack.}]
Dozens of MFA push notifications exhaust user into accidentally approving one; attacker logs in, registers device, creates inbox rule.

\item[\textit{Session Hijacking via Token Theft.}]
Malware extracts browser session tokens; attacker replays sessions without credentials or MFA to access mailbox and SharePoint.

\item[\textit{Help Desk Social Engineering.}]
Attacker impersonates victim to the help desk, convinces IT staff to reset MFA, then enrolls attacker-controlled device.

\item[\textit{LOLBins Living-off-the-Land Attack.}]
Attack chain using legitimate Windows binaries (Excel macro $\to$ PowerShell $\to$ certutil $\to$ mshta) to evade detection.

\item[\textit{Ransomware Deployment Attack.}]
Multi-stage attack starting with RDP brute force; Cobalt Strike beacon, LSASS dump, EDR disabling, VSS deletion, then ransomware deployment across hosts.

\item[\textit{Fileless Malware PowerShell Attack.}]
Teams/Slack link delivers JavaScript loader; PowerShell executes entirely in memory, harvests browser credentials, and exfiltrates via DNS tunneling.

\item[\textit{Personal Email Data Exfiltration.}]
Employee systematically exports sensitive reports and SharePoint files and emails them to a personal Gmail account before deleting sent items.

\item[\textit{Screenshot and Print Exfiltration.}]
Employee captures screenshots from sensitive applications, prints confidential documents, and copies all data to a personal phone via USB.

\end{description}

\subsection*{Tier~3 — Multi-day campaign reconstruction, long-horizon reasoning (26 scenarios)}

\begin{description}[leftmargin=0pt, labelindent=0pt, itemsep=4pt]

\item[\textit{BEC Invoice Diversion --- Payment Redirection Fraud.}]
Attacker compromises a finance-adjacent account, intercepts a legitimate invoice thread, and redirects a wire transfer to an attacker-controlled account across multiple email exchanges.

\item[\textit{MFA and Conditional Access Policy Weakening.}]
Attacker with admin access progressively loosens conditional access policies to exempt compromised accounts from MFA requirements before exfiltrating data.

\item[\textit{Service Principal and App Secret Persistence.}]
Attacker creates service principals with client secrets tied to compromised identities, establishing durable machine-to-machine access that survives password resets.

\item[\textit{Guest User Invitation Abuse for Persistence.}]
Attacker invites attacker-controlled guest accounts with elevated permissions, using legitimate B2B collaboration features to maintain long-term access.

\item[\textit{Federation Trust and SAML Configuration Modification.}]
Attacker with hybrid identity admin rights modifies SAML federation trust to allow forged assertions, enabling persistent authentication bypass.

\item[\textit{Scheduled Task and Registry Run Key Persistence.}]
PowerShell-delivered payload creates scheduled task and Registry Run key for persistence; re-executes multiple times across days while collecting and exfiltrating data.

\item[\textit{DNS Tunneling for Covert C2 and Data Exfiltration.}]
Implant encodes data in DNS TXT queries across multiple days, maintaining a low-volume covert channel that evades volumetric detection.

\item[\textit{Malicious Browser Extension for Session and Credential Theft.}]
Attacker deploys a malicious browser extension that silently harvests session tokens and credentials across multiple web applications over days.

\item[\textit{Group Policy Object Abuse for Mass Payload Deployment.}]
Attacker modifies GPO to deploy a malicious scheduled task across all hosts in a target OU; multiple machines execute the payload after policy refresh.

\item[\textit{AD CS Certificate Abuse (ESC1 / SAN Forgery).}]
Misconfigured certificate template exploited to request a certificate impersonating a Domain Admin; used for persistent Kerberos authentication across multiple systems over days.

\item[\textit{Cloud API Key Creation for Persistence.}]
Attacker creates multiple long-lived API keys across cloud services, establishing persistent access that survives IAM policy changes.

\item[\textit{Cloud Infrastructure Cryptomining Abuse.}]
Attacker provisions high-compute instances across multiple regions for cryptomining, gradually scaling usage to avoid cost-anomaly alerts.

\item[\textit{Mailbox Collection via Platform Export.}]
Attacker with compliance admin credentials creates eDiscovery searches, exports PST files, and exfiltrates them over HTTPS.

\item[\textit{Archive-Based Staging and Exfiltration.}]
Attacker collects files in small batches over multiple days, stages into password-encrypted archives disguised as system backups, and exfiltrates in sized chunks.

\item[\textit{DNS Tunneling Exfiltration.}]
Implant encodes sensitive data in DNS TXT queries across multiple days to stay below volume detection thresholds.

\item[\textit{Double Extortion: Data Theft then Ransomware.}]
Over multiple days, attacker performs reconnaissance, lateral movement, and staged data theft from Finance/HR/Engineering shares before deploying ransomware for dual extortion leverage.

\item[\textit{Backup Destruction Before Ransomware / Wiper.}]
Attacker enumerates all backup infrastructure (VSS, Veeam, cloud vaults) during business hours, then systematically destroys every recovery layer before deploying a wiper.

\item[\textit{Wiper Payload Disguised as Scheduled Maintenance.}]
Destructive payload masquerading as a routine maintenance script is deployed during a legitimate change window to destroy data across multiple systems.

\item[\textit{AiTM Phishing to SharePoint Mass Exfiltration.}]
AiTM session capture followed by MFA device registration, inbox rule creation, email thread hijacking for lateral phishing, and mass SharePoint document exfiltration over HTTPS.

\item[\textit{Device-Code Phishing to DCSync Domain Credential Dump.}]
Device code phishing grants persistent token access; attacker pivots from M365 to on-premises AD via hybrid connector and performs DCSync to extract all domain hashes.

\item[\textit{OAuth Consent Phishing to Slow-Drip Data Exfiltration.}]
Malicious OAuth app established as persistent data pipeline, slowly exfiltrating mailbox and OneDrive content in low-volume batches across weeks.

\item[\textit{Poisoned CI Pipeline to VM Snapshot Exfiltration.}]
Malicious CI workflow dumps cloud credentials, uses them to create VM disk snapshots, and exports them to an attacker-controlled bucket for offline analysis.

\item[\textit{Contractor VPN Compromise to eDiscovery Mass Export.}]
Compromised contractor VPN account used for network recon, lateral movement, AD CS privilege escalation, hybrid identity manipulation, and multi-day eDiscovery mass export.

\item[\textit{USB Malware Propagation.}]
Malware auto-executes from an inserted USB device, replicates to connected drives, establishes C2, and spreads across the network through file share enumeration.

\item[\textit{Mass Download Before Resignation.}]
Resigning employee systematically downloads sensitive data from SharePoint, OneDrive, and network shares across multiple days using a personal sync app before departure.

\item[\textit{Cloud Storage Sync Exfiltration.}]
Employee installs personal cloud sync app and progressively synchronises corporate SharePoint, OneDrive, and network share content to personal storage before sharing externally.

\end{description}
\section{SOC Tool Catalog}
\label{app:soc_tools}

Legion-Bench exposes seven SOC tool categories to investigated agents.
Each category is backed by a named real-platform analogue; tool names
visible to the agent follow the pattern \texttt{<prefix>\_<function>}.
Each entry lists required parameters (\textit{req.}) and optional
parameters (\textit{opt.}) separately.

\subsection*{SIEM\,---\,Splunk Enterprise Security (\texttt{splunk\_*})}
\textit{Data scope:} login, email, file, admin, app, process, MFA, device, print, and network events;
\texttt{employees}, \texttt{devices}, and \texttt{subnets} reference tables.

\begin{description}[leftmargin=0pt,labelindent=0pt,itemsep=3pt]

\item[\texttt{splunk\_run\_query}]
\textit{Req.}~\texttt{query}; \textit{opt.}~\texttt{limit}, \texttt{time\_start}, \texttt{time\_end}.\\
Execute an arbitrary SQL \texttt{SELECT} on any event or reference table.
Supports \texttt{JOIN}s across all ten event tables and the three reference tables.

\item[\texttt{splunk\_search\_app\_events}]
\textit{Opt.}~\texttt{employee\_id}, \texttt{app}, \texttt{action}, \texttt{resource\_type},
\texttt{resource\_name}, \texttt{success}, \texttt{time\_start}, \texttt{time\_end}, \texttt{limit}.\\
Search application activity events (Slack, GitHub, Salesforce, Jira, Okta, \ldots).

\item[\texttt{splunk\_search\_dashboard}]
\textit{Opt.}~\texttt{employee\_id}, \texttt{device\_id}, \texttt{time\_start}, \texttt{time\_end}.\\
Pre-built activity dashboard for a user or device: comprehensive summary across all event types.

\end{description}

\subsection*{EDR\,---\,CrowdStrike Falcon (\texttt{cs\_*})}
\textit{Data scope:} process, device, and network events; login events; employee and device tables.

\begin{description}[leftmargin=0pt,labelindent=0pt,itemsep=3pt]

\item[\texttt{cs\_get\_process\_tree}]
\textit{Opt.}~\texttt{device\_id}, \texttt{employee\_id}, \texttt{process\_name},
\texttt{time\_start}, \texttt{time\_end}.\\
Parent-child process execution tree for a device or user.

\item[\texttt{cs\_search\_process\_by\_hash}]
\textit{Req.}~\texttt{hash}.\\
Find all executions of a binary by SHA-256 hash, showing which hosts ran it and when.

\item[\texttt{cs\_search\_process\_by\_command}]
\textit{Req.}~\texttt{command\_pattern}; \textit{opt.}~\texttt{time\_start}, \texttt{time\_end}.\\
Find processes by command-line substring --- useful for detecting encoded payloads or suspicious arguments.

\item[\texttt{cs\_get\_host\_timeline}]
\textit{Req.}~\texttt{device\_id}; \textit{opt.}~\texttt{time\_start}, \texttt{time\_end}, \texttt{limit}.\\
Chronological endpoint timeline: processes, USB connections, and software installs.

\item[\texttt{cs\_get\_device\_events}]
\textit{Opt.}~\texttt{employee\_id}, \texttt{device\_id}, \texttt{action},
\texttt{time\_start}, \texttt{time\_end}.\\
USB connections, Bluetooth events, and software installs/uninstalls.
\texttt{action}: \texttt{usb\_connected}, \texttt{usb\_disconnected}, \texttt{software\_install},
\texttt{software\_uninstall}, \texttt{bluetooth\_connect}, \texttt{bluetooth\_disconnect}.

\item[\texttt{cs\_detect\_suspicious\_processes}]
\textit{Opt.}~\texttt{suspicious\_names}, \texttt{time\_start}, \texttt{time\_end}.\\
Detect known LOLBins and hacking tools. A custom name list overrides the built-in default.

\item[\texttt{cs\_get\_network\_events}]
\textit{Opt.}~\texttt{source\_ip}, \texttt{destination\_ip}, \texttt{action},
\texttt{time\_start}, \texttt{time\_end}, \texttt{limit}.\\
Raw network events (firewall logs, connection records).

\item[\texttt{cs\_search\_network\_events}]
\textit{Opt.}~\texttt{ip}, \texttt{url}, \texttt{time\_start}, \texttt{time\_end}, \texttt{limit}.\\
Search network events by IP or domain pattern, matching both source and destination fields.

\end{description}

\subsection*{Identity\,---\,Microsoft Entra ID (\texttt{entra\_*})}
\textit{Data scope:} login events, MFA events, admin actions, employee and device records, group memberships.

\begin{description}[leftmargin=0pt,labelindent=0pt,itemsep=3pt]

\item[\texttt{entra\_lookup\_user}]
\textit{Req.}~\texttt{user\_ref} (ID, email, or display name).\\
Resolve a user to a full profile including authorisations and linked accounts.

\item[\texttt{entra\_sign\_in\_logs}]
\textit{Opt.}~\texttt{employee\_id}, \texttt{email}, \texttt{success},
\texttt{time\_start}, \texttt{time\_end}, \texttt{limit}.\\
Authentication events with IP address, MFA status, geography, and accessed applications.

\item[\texttt{entra\_mfa\_activity}]
\textit{Opt.}~\texttt{employee\_id}, \texttt{action}, \texttt{time\_start}, \texttt{time\_end}.\\
MFA push notifications, approvals/denials, and device registrations/removals.
\texttt{action}: \texttt{push\_sent}, \texttt{push\_approved}, \texttt{push\_denied},
\texttt{device\_registered}, \texttt{device\_removed}, \texttt{code\_entered}, \texttt{code\_failed}.

\item[\texttt{entra\_get\_user\_groups}]
\textit{Req.}~\texttt{user\_ref}.\\
All security and distribution groups a user belongs to, with each group's authorisations.

\item[\texttt{entra\_get\_group\_members}]
\textit{Req.}~\texttt{group\_ref}.\\
All members of a security or distribution group.

\item[\texttt{entra\_user\_audit\_log}]
\textit{Opt.}~\texttt{employee\_id}, \texttt{action}, \texttt{time\_start},
\texttt{time\_end}, \texttt{limit}.\\
Audit trail of account changes: group modifications, role assignments, password resets.

\item[\texttt{entra\_user\_access\_review}]
\textit{Req.}~\texttt{user\_ref}.\\
Comprehensive access review: all groups, authorisations, devices, and recent admin changes.

\item[\texttt{entra\_risky\_users}]
\textit{Opt.}~\texttt{time\_start}, \texttt{time\_end}.\\
Users with risky sign-in patterns: failed logins, MFA denials, and impossible travel indicators.

\end{description}

\subsection*{Email Security\,---\,Microsoft Defender for Office\,365 (\texttt{defender\_*})}
\textit{Data scope:} email events, employee records, mailboxes and distribution lists.

\begin{description}[leftmargin=0pt,labelindent=0pt,itemsep=3pt]

\item[\texttt{defender\_search\_emails}]
\textit{Opt.}~\texttt{sender}, \texttt{subject\_contains}, \texttt{recipient}, \texttt{ip},
\texttt{url}, \texttt{filename}, \texttt{time\_start}, \texttt{time\_end}, \texttt{limit}.\\
Search the message trace by any combination of indicators; filters combine with AND logic.

\item[\texttt{defender\_get\_email\_attachments}]
\textit{Opt.}~\texttt{sender}, \texttt{filename\_contains}, \texttt{file\_type},
\texttt{time\_start}, \texttt{time\_end}.\\
Find emails carrying specific attachments, filtered by sender, filename substring, or extension.

\item[\texttt{defender\_get\_sender\_summary}]
\textit{Req.}~\texttt{sender}.\\
Profile a sender: message count, recipients reached, subject distribution, and attachment statistics.

\item[\texttt{defender\_find\_phishing\_indicators}]
\textit{Opt.}~\texttt{sender\_domain}, \texttt{time\_start}, \texttt{time\_end}.\\
Identify external emails with phishing signals: embedded URLs, executable attachments, or urgency subjects.

\end{description}

\subsection*{DLP / IRM\,---\,Microsoft Purview (\texttt{purview\_*})}
\textit{Data scope:} file events, print events, email events, employee records.

\begin{description}[leftmargin=0pt,labelindent=0pt,itemsep=3pt]

\item[\texttt{purview\_get\_dlp\_alerts}]
\textit{Opt.}~\texttt{employee\_id}, \texttt{policy\_name}, \texttt{time\_start},
\texttt{time\_end}, \texttt{limit}.\\
DLP policy violation alerts: sensitive-data sharing, unauthorised transfers, and classification breaches.

\item[\texttt{purview\_get\_irm\_alerts}]
\textit{Opt.}~\texttt{employee\_id}, \texttt{policy\_name}, \texttt{time\_start},
\texttt{time\_end}, \texttt{limit}.\\
Insider Risk Management alerts: bulk downloads, unusual file-access patterns, print anomalies, and exfiltration indicators.

\end{description}

\subsection*{Threat Intelligence\,---\,VirusTotal / AbuseIPDB (\texttt{ti\_*})}
\textit{Data scope:} attack ground-truth IOC feeds, process events, email events, login events.

\begin{description}[leftmargin=0pt,labelindent=0pt,itemsep=3pt]

\item[\texttt{ti\_lookup\_domain}]
\textit{Req.}~\texttt{domain}.\\
Check a domain against threat feeds. Returns verdict (malicious / unknown) and associated campaigns.

\item[\texttt{ti\_lookup\_ip}]
\textit{Req.}~\texttt{ip\_address}.\\
Check an IP address. Returns verdict and associated attack campaigns.

\item[\texttt{ti\_lookup\_hash}]
\textit{Req.}~\texttt{file\_hash}.\\
Check a SHA-256 (or other) hash. Returns verdict, first-seen date, and affected host count.

\item[\texttt{ti\_lookup\_url}]
\textit{Req.}~\texttt{url}.\\
Check a URL. Returns verdict, associated attacks, and email delivery count.

\item[\texttt{ti\_lookup\_email\_address}]
\textit{Req.}~\texttt{email}.\\
Check an email address: internal flag, known attack associations, and total outbound message count.

\item[\texttt{ti\_lookup\_filename}]
\textit{Req.}~\texttt{filename}.\\
Check a filename against known malicious names in attack campaigns, process events, and email attachments.

\item[\texttt{ti\_get\_known\_iocs}]
\textit{Opt.}~\texttt{ioc\_type} (\texttt{domain}, \texttt{ip}, \texttt{hash},
\texttt{url}, \texttt{email}, \texttt{filename}).\\
List all IOCs from threat feeds, optionally filtered by type.

\end{description}

\subsection*{CNAPP\,---\,Wiz (\texttt{cnapp\_*})}
\textit{Data scope:} cloud instances, containers, vulnerabilities, misconfigurations, threats,
and API audit logs (AWS CloudTrail, Azure Activity Log, GCP Audit Log).

\begin{description}[leftmargin=0pt,labelindent=0pt,itemsep=3pt]

\item[\texttt{cnapp\_search\_resource}]
\textit{Opt.}~\texttt{resource\_id}, \texttt{identity}, \texttt{provider}, \texttt{resource\_type}.\\
Discover cloud resources or IAM identities with network configuration, permissions, and findings.

\item[\texttt{cnapp\_search\_vulnerabilities}]
\textit{Opt.}~\texttt{resource\_id}, \texttt{cve\_id}, \texttt{severity}, \texttt{provider}, \texttt{limit}.\\
CWPP vulnerability findings by CVE, severity, or resource.
\texttt{severity}: \texttt{critical}, \texttt{high}, \texttt{medium}, \texttt{low}.

\item[\texttt{cnapp\_search\_misconfigurations}]
\textit{Opt.}~\texttt{resource\_id}, \texttt{category}, \texttt{severity}, \texttt{provider}, \texttt{limit}.\\
CSPM findings: open security groups, public buckets, missing encryption, logging gaps, and more.
\texttt{category}: \texttt{networking}, \texttt{encryption}, \texttt{access\_control}, \texttt{logging}.

\item[\texttt{cnapp\_get\_threat\_details}]
\textit{Opt.}~\texttt{resource\_id}, \texttt{identity}, \texttt{threat\_type},
\texttt{severity}, \texttt{limit}.\\
Cloud threat detections: unauthorized access, cryptomining, data exfiltration, and related attacks.

\item[\texttt{cnapp\_search\_cloud\_events}]
\textit{Opt.}~\texttt{instance\_id}, \texttt{identity}, \texttt{event\_name},
\texttt{event\_source}, \texttt{event\_category}, \texttt{provider}, \texttt{source\_ip},
\texttt{response\_status}, \texttt{read\_only}, \texttt{time\_start}, \texttt{time\_end}, \texttt{limit}.\\
Query cloud API audit logs across AWS, Azure, and GCP.
\texttt{event\_category}: \texttt{management}, \texttt{data}, \texttt{network}, \texttt{auth}, \texttt{monitoring}.
\texttt{response\_status}: \texttt{Success}, \texttt{Failed}, \texttt{AccessDenied}.

\item[\texttt{cnapp\_get\_cloud\_identity\_activity}]
\textit{Req.}~\texttt{identity}; \textit{opt.}~\texttt{time\_start}, \texttt{time\_end}, \texttt{limit}.\\
All cloud API calls made by a specific IAM identity --- useful for investigating compromised service accounts.

\item[\texttt{cnapp\_get\_cloud\_event\_summary}]
\textit{Opt.}~\texttt{instance\_id}, \texttt{identity}, \texttt{provider},
\texttt{time\_start}, \texttt{time\_end}.\\
Cloud API calls aggregated by event name with counts and time ranges, enabling anomaly detection.

\end{description}

\section{Agent and Evaluator Prompts}
\label{app:prompts}

All agents share a common \textbf{Report Format} template and
\textbf{Employee-ID note} injected into their respective system prompts.
These are shown first, followed by the architecture-specific prompts
and finally the LLM judge prompt used for Findings Recall evaluation.

\subsection*{Shared: Report Format (injected into all agent system prompts)}
\label{app:report_format}
\begin{small}
\begin{verbatim}
**Executive Summary**: One-paragraph overview of what happened,
the severity, and immediate risk.

**Findings** (focus on the essential elements that define this attack
— be specific and include evidence):
- The specific attack technique or vector used (what makes this attack
  distinctive)
- Key attacker actions that confirm the attack (e.g., impossible travel,
  credential replay, privilege escalation)
- Persistence or defense evasion mechanisms (e.g., MFA registration,
  inbox rules, scheduled tasks, service principals)
- Ultimate impact — what data, systems, or access was compromised
- Any lateral movement to other accounts, hosts, or cloud resources

Prioritize the core narrative of the attack over exhaustive enumeration
of every step. Focus on the critical evidence that proves the attack
happened, identifies the technique, and defines its scope and impact.

**Indicators of Compromise (IOCs)** -- list each IOC belonging to the
attacker on its own line. The different types of IOCs are: IPs, domains,
URLs, emails, hashes, and filenames. Just list the IOCs, no other text.

**Severity Assessment**: Your assessment of the incident severity.
\end{verbatim}
\end{small}

\noindent\textit{Employee-ID note (appended to every system prompt):} ``Employee IDs are different from their email addresses. Almost all functions require employee IDs and will not work with email addresses. The first step needs to be to get the employee ID from the email address.''

\subsection*{Direct Agent}
\begin{small}
\begin{verbatim}
You are an expert SOC (Security Operations Center) analyst conducting
a security investigation.

## Your Task

You have been given a security alert to investigate. Use the available
tools to gather evidence, analyze the situation, and produce a
comprehensive investigation report.

## Tool Usage

- You have access to SOC tools (SIEM, EDR, identity, email gateway,
  DLP, threat intel, CNAPP).
- Each tool call returns structured data (lists of events, user
  records, etc.).
- Call multiple tools per turn when the queries are independent.

## Final Report

When you have gathered enough evidence, provide your final investigation
report with this structure:

[REPORT FORMAT]

Do NOT call any tools when providing your final report — just output
the report text.
\end{verbatim}
\end{small}

\subsection*{ReAct Agent}
\begin{small}
\begin{verbatim}
You are an expert SOC (Security Operations Center) analyst conducting
a security investigation. You follow a strict Thought-Action-Observation
loop.

## Your Methodology

On EVERY turn, you MUST follow this exact pattern:

**Thought**: Reason about the current state of the investigation.
What do you know so far? What hypotheses are you testing? What
information do you still need?

**Action**: Based on your reasoning, call one or more tools to gather
evidence. Call multiple tools per turn when the queries are independent.

The system will return **Observation** results from your tool calls,
and then you repeat the cycle with a new Thought.

## Important Rules

1. ALWAYS start your response with "**Thought**:" followed by your
   reasoning.
2. After your thought, make your tool calls (the Action).
3. Continue this Thought-Action-Observation cycle until you have
   enough evidence.
4. When ready to conclude, output one final "**Thought**:" summarizing
   your findings, then provide your report WITHOUT any tool calls.

[REPORT FORMAT]
\end{verbatim}
\end{small}

\subsection*{SPA (See--Plan--Act) Agent}
\begin{small}
\begin{verbatim}
You are an expert SOC analyst. You follow a strict Plan-then-Act
methodology.

## Your Methodology

### Planning Phase
- Analyze the alert, indicators, and any evidence gathered so far.
- Create a structured investigation plan listing the specific queries
  you will run and why.
- Do NOT attempt any tool calls during planning — just output your
  plan as text.

### Execution Phase
- Execute your plan by calling the available tools.
- Call multiple tools per turn when the queries are independent.
- If your findings reveal that your plan needs revision (e.g.,
  unexpected results, new leads, dead ends), call the `request_replan`
  tool to return to the planning phase.
- When you have gathered enough evidence, provide your final report
  without any tool calls.

[REPORT FORMAT]

--- [Execute prompt, injected after plan is produced] ---

Good. Now execute your investigation plan using the available tools.
Call multiple tools per turn when the queries are independent. If at
any point your findings require a change in investigation direction,
call the `request_replan` tool to create a revised plan. When you
have gathered enough evidence, provide your final report without any
tool calls.

--- [Replan prompt, injected on `request_replan` call] ---

You requested to revise your investigation plan. Reason: {reason}

Based on everything you've learned so far, create a new investigation
plan. Consider:
- What have you confirmed so far?
- What questions remain unanswered?
- What new leads or directions have emerged?
- What specific queries will you run next and why?

Output only your revised plan — do not call any tools.
\end{verbatim}
\end{small}

\subsection*{Reflexion Agent}

\paragraph{Pass~1 Investigator (same as Direct; produces draft report).}

\paragraph{Critic system prompt.}
\begin{small}
\begin{verbatim}
You are a senior SOC analyst reviewing a junior analyst's investigation
report. Your job is to identify gaps, errors, and areas that need
deeper investigation.

Be specific and actionable in your critique. For each gap, explain
exactly what additional queries or analysis should be performed.
\end{verbatim}
\end{small}

\paragraph{Critic user prompt.}
\begin{small}
\begin{verbatim}
## Original Alert
{alert_summary}

## Draft Investigation Report
{draft_report}

## Review Instructions

Critically review this investigation report. Focus on whether it
captures the essential elements of the attack — not every minor detail,
but the core findings that matter. Identify:

1. **Attack technique unclear**: Is the specific attack vector clearly
   identified with evidence?
2. **Missing key attacker actions**: Were critical pivot points
   (e.g., credential replay, privilege escalation, persistence setup)
   investigated and confirmed?
3. **Impact not established**: Is it clear what data, systems, or
   access was ultimately compromised?
4. **Unverified claims**: What conclusions lack supporting evidence
   from tool queries?
5. **Missing IOC categories**: Were all IOC types checked (IPs,
   domains, hashes, emails)?

Focus your critique on gaps in the core attack narrative — the
technique, the key indicators, persistence mechanisms, and impact.
Do not flag missing details that are supplementary (e.g., every
intermediate recon step or every file accessed).

For each gap, provide a specific recommendation for what the analyst
should investigate next.
\end{verbatim}
\end{small}

\paragraph{Pass~2 system prompt.}
\begin{small}
\begin{verbatim}
You are an expert SOC analyst conducting a follow-up investigation to
address gaps identified in a peer review of your initial findings.

## Your Task

A senior analyst has reviewed your draft report and identified gaps.
Use the available tools to address each gap, then produce an improved
final report that incorporates both your original findings and the
new evidence.

[REPORT FORMAT]
\end{verbatim}
\end{small}

\paragraph{Pass~2 user prompt.}
\begin{small}
\begin{verbatim}
## Your Draft Report
{draft_report}

## Senior Analyst Critique
{critique}

Address the gaps identified above. Use the available tools to gather
additional evidence, then produce an improved final report.
\end{verbatim}
\end{small}

\subsection*{Multi-Agent Architecture}

\paragraph{Coordinator triage system prompt.}
\begin{small}
\begin{verbatim}
You are a senior SOC coordinator. Your role is to triage security
alerts and assign investigation tasks to domain specialists.

You do NOT have access to any tools. Your job is to analyze the alert
and create a structured task assignment for each specialist.
\end{verbatim}
\end{small}

\paragraph{Coordinator triage user prompt.}
\begin{small}
\begin{verbatim}
## Alert: {alert_title}

**Severity**: {severity}
**Source**: {source}

### Description
{description}

### Initial Indicators
{indicators}

### Available Specialists
{specialists}

## Instructions

Analyze this alert and create specific investigation tasks for each
available specialist. For each specialist, provide:
1. What they should investigate
2. Specific queries or checks to perform
3. What evidence to look for
\end{verbatim}
\end{small}

\paragraph{Specialist system prompt (instantiated per domain).}
\begin{small}
\begin{verbatim}
You are a specialist SOC analyst in the **{role}** domain. You are
part of a team investigating a security alert. Your coordinator has
assigned you specific tasks.

## Your Scope

You only have access to tools in your domain. Focus on your assigned
tasks and produce a concise findings report for the coordinator.

## Output

When done, provide a structured summary of your findings, focusing on:
- The core attack technique or attacker actions you identified
  (with evidence)
- Any persistence or defense evasion mechanisms discovered
- IOCs found (IPs, domains, hashes, emails, URLs)
- Risk assessment for your domain
- Any leads that other specialists should follow up on

Focus on the critical evidence that defines the attack — not every
minor event, but the key findings that prove what happened and its
impact.
\end{verbatim}
\end{small}

\paragraph{Coordinator synthesis system prompt.}
\begin{small}
\begin{verbatim}
You are a senior SOC coordinator synthesizing investigation findings
from your specialist team into a final, comprehensive investigation
report.

You do NOT have access to any tools. Synthesize all specialist
findings into a single report.

[REPORT FORMAT]
\end{verbatim}
\end{small}

\paragraph{Coordinator synthesis user prompt.}
\begin{small}
\begin{verbatim}
## Original Alert
{alert_summary}

## Specialist Reports
{specialist_reports}

## Instructions

Synthesize all specialist findings into a single comprehensive
investigation report. Resolve any contradictions, identify patterns
across domains, and provide a unified timeline and assessment.

Focus the report on the essential attack narrative: the specific
technique used, the key attacker actions that confirm the attack,
persistence/evasion mechanisms, and the ultimate impact. Prioritize
critical evidence over exhaustive enumeration of every step.
\end{verbatim}
\end{small}

\subsection*{Findings Recall Evaluator (LLM Judge)}
\label{app:judge_prompt}

The judge is called with \texttt{temperature=0.0} using \texttt{gpt-4o-mini}
and is instructed to return strict JSON (no markdown fences) containing (found / not found, supporting quote, confidence).

\paragraph{System prompt.}
\begin{small}
\begin{verbatim}
You are a strict evaluation judge for a SOC (Security Operations Center)
investigation benchmark.

You will be given:
1. An investigation report written by an AI agent.
2. A list of expected findings that the agent should have identified.

For each expected finding, determine whether the report contains
SPECIFIC, CONCRETE information that substantively demonstrates the
agent discovered this finding through its investigation. Apply these
strict rules:

MATCHING RULES:
- The report must show the agent actually IDENTIFIED and INVESTIGATED
  the finding, not merely restated the alert description or ticket
  information.
- Semantic equivalence is acceptable — the match does NOT need to be
  verbatim. For example, "bulk download of files" matches "mass file
  download".
- Expected findings describe the essential, high-level elements of the
  attack (technique, key actions, persistence, impact). Match
  generously on scope — if the expected finding is broad (e.g., "data
  exfiltrated over C2"), the report satisfies it by showing ANY
  specific evidence of that activity (e.g., mentioning the C2 IP and
  outbound transfer).
- A finding is FOUND only if the report provides specific evidence or
  details about it (e.g., specific IPs, timestamps, user actions, tool
  outputs).
- A finding is NOT FOUND if:
  * The report only vaguely alludes to it without specifics
  * The report merely repeats information from the initial alert/ticket
    without adding investigation results
  * The information is mentioned in a recommendation section but was
    never actually confirmed
  * The report speculates about it without evidence from tool queries

Return a JSON object with a single key "results" containing an array
with one entry per expected finding, in the same order. Each entry
must have:
- "index": the 0-based index of the finding
- "found": true/false
- "evidence": a brief quote or paraphrase from the report that
  supports your decision (or "not found in report" if not found)
- "confidence": a float 0.0-1.0 indicating your confidence

Example output:
{"results": [{"index": 0, "found": true,
  "evidence": "Report states: Query returned 2847 file download events
  from SharePoint between March 10-12", "confidence": 0.95}]}

IMPORTANT: Return ONLY valid JSON, no markdown fences or extra text.
Be strict — when in doubt, mark as not found.
\end{verbatim}
\end{small}

\paragraph{User prompt.}
\begin{small}
\begin{verbatim}
## Investigation Report

{report}

## Expected Findings

{numbered_list_of_expected_findings}
\end{verbatim}
\end{small}

\section{Golden Workflow Baseline}
\label{app:golden_workflow}

The \emph{Golden Workflow} is a deterministic playbook included in
\textsc{Legion Security Bench} as a reproducible upper-bound baseline.
Unlike the five LLM-driven agents (Direct, ReAct, SPA, Reflexion,
Multi-Agent), which choose every tool call dynamically, the Golden
Workflow executes a fixed, pre-specified sequence of SOC tool calls
derived from expert-authored playbooks, then issues a \emph{single} LLM
call to synthesise the collected evidence into a final report.
Because the tool call sequence is fully determined by the alert type and
the content of earlier results, two runs on the same alert and database
always produce identical tool calls, making the baseline reproducible
and free of sampling variance.

\subsection*{Architecture Overview}

Execution proceeds in two stages:

\begin{enumerate}
  \item \textbf{Deterministic execution.}
    The alert is routed to one of four typed workflows
    (AccountTakeover, Phishing, Malware, or DLP).
    Each workflow is a list of \texttt{WorkflowStep} objects.
    Every step specifies a tool name and a \emph{builder function} that
    maps the current \texttt{ExecutionContext} --- the mutable state
    accumulating all prior tool call results --- to a list of argument
    dicts.
    Steps that fan out (e.g., one TI lookup per suspicious IP observed in
    sign-in logs) produce multiple parallel tool calls; steps whose
    builder returns an empty list are silently skipped.
    No LLM is involved in deciding which tool to call or what arguments
    to pass.

  \item \textbf{Single LLM synthesis.}
    Once all steps complete, a single LLM call (using the same model as
    the evaluated agents) receives the full evidence corpus and produces
    the investigation report in the shared report format
    (\S\ref{app:prompts}).
\end{enumerate}

\noindent The \texttt{ExecutionContext} provides typed extraction helpers
(\texttt{collect\_ips}, \texttt{collect\_domains}, \texttt{collect\_urls},
\texttt{collect\_hashes}, \texttt{collect\_external\_senders},
\texttt{collect\_filenames}) that parse prior tool results to derive the
parameters for downstream steps, implementing the data-dependency graph
of each playbook without any LLM reasoning.

\subsection*{Alert Routing}

Alerts are routed deterministically by the \texttt{pick\_golden\_workflow}
function, which maps the alert's \texttt{source} field to a workflow:

\begin{center}
\begin{tabular}{ll}
  \toprule
  \textbf{Alert Source} & \textbf{Workflow} \\
  \midrule
  \texttt{purview}        & DLP \\
  \texttt{crowdstrike}    & Malware \\
  \texttt{defender\_o365} & Phishing \\
  \texttt{entra\_id}      & AccountTakeover \\
  \texttt{splunk\_es}     & AccountTakeover (default) or Malware \\
                          & \quad(title prefix match for supply-chain alerts) \\
  \bottomrule
\end{tabular}
\end{center}

\noindent Routing achieves 100\% alignment with the intended workflow on
the 98 scenario tickets in our benchmark fixtures (verified against
ground-truth attack types).
A small number of structurally ambiguous alerts --- such as a personal
email exfiltration ticket fired by \texttt{defender\_o365} --- are
intentionally routed by ticket framing rather than underlying attack
type, mirroring real analyst triage practice.

\subsection*{AccountTakeover Workflow (13 Steps)}

Designed for identity-compromise alerts from Entra~ID or Splunk~ES.
Phases map to the pre-Verdict nodes of the AccountTakeover SOC playbook.

\begin{description}[leftmargin=0pt,labelindent=0pt,itemsep=3pt]
  \item[\textsc{User Context} --- \texttt{entra\_lookup\_user}]
    Resolve the alerted email to a full user profile and authorisation set.
  \item[\textsc{User Groups} --- \texttt{entra\_get\_user\_groups}]
    Enumerate security group memberships to establish blast radius.
  \item[\textsc{Sign-in Logs} --- \texttt{entra\_sign\_in\_logs}]
    Retrieve authentication events in a $\pm$2-day window around the alert
    timestamp (up to 200 events).
  \item[\textsc{Audit Log} --- \texttt{entra\_user\_audit\_log}]
    Fetch the account-change audit trail: password resets, MFA modifications,
    group changes, and delegation rules.
  \item[\textsc{MFA Activity} --- \texttt{entra\_mfa\_activity}]
    Retrieve push notifications, approvals/denials, and device
    registration events for the same window.
  \item[\textsc{Risk Detections} --- \texttt{entra\_risky\_users}]
    Query risky sign-in patterns: impossible travel, anomalous failure
    bursts, and leaked credential indicators.
  \item[\textsc{IP Threat Intel} --- \texttt{ti\_lookup\_ip} (fan-out)]
    For each distinct external IP observed in sign-in logs, submit a
    threat-feed lookup.  Multiple playbook nodes (IP\_REP, IP\_GEO,
    IP\_VPN, IP\_TI\_FEED) collapse to a single call per IP.
  \item[\textsc{IP Org History} --- \texttt{splunk\_run\_query}]
    Cross-reference suspicious IPs against the full login table to
    identify other accounts targeted from the same source addresses.
  \item[\textsc{Adversary Hunting} --- \texttt{splunk\_search\_dashboard}]
    Broad activity dashboard for the victim to surface anomalies not
    captured by prior targeted queries.
  \item[\textsc{Endpoint Device Events} --- \texttt{cs\_get\_device\_events}]
    Check for USB insertions, Bluetooth connections, and software installs
    on the victim's enrolled device.
  \item[\textsc{Process Tree} --- \texttt{cs\_get\_process\_tree}]
    Inspect endpoint process execution for post-compromise tooling or
    persistence mechanisms.
  \item[\textsc{Network Corroboration} --- \texttt{cs\_search\_network\_events} (fan-out)]
    For each suspicious IP, query whether it communicated with the
    victim's endpoint.
  \item[\textsc{Phishing Correlation} --- \texttt{defender\_search\_emails}]
    Search the mail gateway for phishing emails delivered to the victim
    before the anomalous sign-in, linking the potential initial access vector.
\end{description}

\subsection*{Phishing Workflow (19 Steps)}

Designed for email-gateway alerts from Microsoft Defender for Office~365.

\begin{description}[leftmargin=0pt,labelindent=0pt,itemsep=3pt]
  \item[\textsc{Recipient Context} --- \texttt{entra\_lookup\_user}]
    Resolve recipient identity and access profile.
  \item[\textsc{Delivery Status} --- \texttt{defender\_search\_emails}]
    Retrieve all emails delivered to the recipient in a $\pm$2-day window,
    extracting sender addresses, URLs, and attachment metadata.
  \item[\textsc{Sender Profile} --- \texttt{defender\_get\_sender\_summary} (fan-out)]
    Profile each external sender: message volume, recipient spread, and
    attachment statistics.
  \item[\textsc{Sender Domain History} --- \texttt{defender\_find\_phishing\_indicators} (fan-out)]
    Query the 30-day phishing-indicator history for each external sender domain.
  \item[\textsc{Sender Email TI} --- \texttt{ti\_lookup\_email\_address} (fan-out)]
    Threat-feed lookup on each external sender address.
  \item[\textsc{Sender Domain TI} --- \texttt{ti\_lookup\_domain} (fan-out)]
    Threat-feed lookup on the domain of each external sender.
  \item[\textsc{URL TI} --- \texttt{ti\_lookup\_url} (fan-out)]
    Threat-feed lookup on each URL extracted from delivered emails.
  \item[\textsc{URL Domain TI} --- \texttt{ti\_lookup\_domain} (fan-out)]
    Threat-feed lookup on the hostname of each embedded URL, providing
    the URL\_CATEGORY signal present in the playbook.
  \item[\textsc{Attachment Metadata} --- \texttt{defender\_get\_email\_attachments} (fan-out)]
    Retrieve detailed attachment metadata per external sender.
  \item[\textsc{Attachment Hash TI} --- \texttt{ti\_lookup\_hash} (fan-out)]
    Threat-feed lookup on SHA-256 hashes of all retrieved attachments.
  \item[\textsc{Attachment Filename TI} --- \texttt{ti\_lookup\_filename} (fan-out)]
    Check attachment filenames against known malicious filename feeds.
  \item[\textsc{Campaign Scope} --- \texttt{defender\_search\_emails} (fan-out)]
    Search the last 7~days for all recipients of the same sender,
    scoping the phishing campaign.
  \item[\textsc{Sign-in Logs} --- \texttt{entra\_sign\_in\_logs}]
    Detect post-click credential submission: foreign IP login shortly
    after email delivery.
  \item[\textsc{Audit Log} --- \texttt{entra\_user\_audit\_log}]
    Inbox rules, MFA changes, and delegation modifications indicating
    post-compromise persistence.
  \item[\textsc{MFA Activity} --- \texttt{entra\_mfa\_activity}]
    MFA push history to detect fatigue attacks or unauthorised device enrolment.
  \item[\textsc{Post-click IP TI} --- \texttt{ti\_lookup\_ip} (fan-out)]
    Threat-feed lookup on each IP observed in post-delivery sign-in logs.
  \item[\textsc{Endpoint Processes} --- \texttt{cs\_get\_process\_tree}]
    Process tree to detect payload execution following a malicious attachment open.
  \item[\textsc{Endpoint Device Events} --- \texttt{cs\_get\_device\_events}]
    USB and device events on the victim endpoint.
  \item[\textsc{User Activity Dashboard} --- \texttt{splunk\_search\_dashboard}]
    Comprehensive cross-source activity summary to surface any signals
    missed by the targeted steps.
\end{description}

\subsection*{Malware Workflow (14 Steps)}

Designed for EDR-sourced alerts from CrowdStrike Falcon.
The workflow pivots on endpoint telemetry, establishing the process
footprint before enriching discovered artefacts with threat intelligence
and correlating across identity, network, and email sources.
Sandbox-based nodes (HASH\_SANDBOX, URL\_SANDBOX) and the CMD\_ANALYZE
node are skipped, as no detonation service is available and LLM calls
during execution are explicitly excluded.

\begin{description}[leftmargin=0pt,labelindent=0pt,itemsep=3pt]
  \item[\textsc{User Context} --- \texttt{entra\_lookup\_user}]
    Phase~1: directory profile of the affected user to establish identity
    and authorisation context.
  \item[\textsc{User Sign-ins} --- \texttt{entra\_sign\_in\_logs}]
    Phase~1: recent authentication logs for host attribution and
    identification of foreign source IPs.
  \item[\textsc{User Audit} --- \texttt{entra\_user\_audit\_log}]
    Phase~1: post-compromise account changes (password resets, MFA
    modifications, new delegations).
  \item[\textsc{Endpoint Process Tree} --- \texttt{cs\_get\_process\_tree}]
    Phase~3: full parent-child process execution tree on the user's
    device across the alert window.
  \item[\textsc{Endpoint Device Events} --- \texttt{cs\_get\_device\_events}]
    Phase~3: USB connections, software installs, and removable-media
    events on the endpoint.
  \item[\textsc{LOLBin Scan} --- \texttt{cs\_detect\_suspicious\_processes}]
    Phase~3: sweep for known living-off-the-land binaries and offensive
    tools (\texttt{powershell}, \texttt{mshta}, \texttt{certutil},
    \texttt{mimikatz}, \texttt{rubeus}, etc.).
  \item[\textsc{Hash TI} --- \texttt{ti\_lookup\_hash} (fan-out)]
    Phase~2: threat-feed lookup per distinct binary hash observed in
    process tree or device events.  Collapses playbook nodes HASH\_REP,
    HASH\_INTEL, HASH\_OSINT, and HASH\_SIGNED into a single call per hash.
  \item[\textsc{Hash Prevalence} --- \texttt{cs\_search\_process\_by\_hash} (fan-out)]
    Phase~2: check whether the same hash has executed on other endpoints
    in the environment (HASH\_ENV node).
  \item[\textsc{IP Threat Intel} --- \texttt{ti\_lookup\_ip} (fan-out)]
    Phase~2: threat-feed lookup per external IP extracted from process
    tree, device, or sign-in events (covers IP\_REP, IP\_ABUSE, IP\_GEO).
  \item[\textsc{Network Corroboration} --- \texttt{cs\_search\_network\_events} (fan-out)]
    Phase~3: network telemetry per suspicious IP to detect C2
    communication and lateral movement.
  \item[\textsc{Domain TI} --- \texttt{ti\_lookup\_domain} (fan-out)]
    Phase~2: threat-feed lookup per distinct domain from emails or
    network events (covers URL\_REP, URL\_CATEGORY).
  \item[\textsc{User Activity Dashboard} --- \texttt{splunk\_search\_dashboard}]
    Phase~3: SIEM activity dashboard across login, file, and application
    event types to surface anomalies not captured by targeted queries.
  \item[\textsc{Email Trail} --- \texttt{defender\_search\_emails}]
    Phase~3: search the mail gateway for emails delivered to the victim
    that may have carried the initial malware payload.
  \item[\textsc{Filename TI} --- \texttt{ti\_lookup\_filename} (fan-out)]
    Phase~3: threat-feed lookup per attachment filename from email lures
    associated with the incident.
\end{description}

\subsection*{DLP Workflow (15 Steps)}

Designed for data-loss prevention alerts from Microsoft Purview.
The workflow begins with user context and a 90-day DLP/IRM history to
establish whether the event is part of a pattern, then inspects all
plausible exfiltration channels (file access, outbound email, print,
cloud sharing, removable media) before enriching external destinations
with threat intelligence.
Nodes without a registry equivalent are skipped: trusted third-party
GRC verification, OSINT-based destination reputation, file-content
encryption analysis, and CASB cloud-sharing scope.

\begin{description}[leftmargin=0pt,labelindent=0pt,itemsep=3pt]
  \item[\textsc{User Context} --- \texttt{entra\_lookup\_user}]
    Phase~2A: directory profile of the user triggering the DLP alert.
  \item[\textsc{User Groups} --- \texttt{entra\_get\_user\_groups}]
    Phase~2A: group memberships to determine authorised data scope and
    blast radius.
  \item[\textsc{DLP History} --- \texttt{purview\_get\_dlp\_alerts}]
    Phase~2A: 90-day DLP alert history to detect repeat-offender patterns
    (Background Analytics node).
  \item[\textsc{IRM Alerts} --- \texttt{purview\_get\_irm\_alerts}]
    Phase~2A: Insider Risk Management signals over 90~days — bulk
    download indicators, leaver-risk flags, and anomalous access patterns.
  \item[\textsc{User Access Review} --- \texttt{entra\_user\_access\_review}]
    Phase~2A: comprehensive access review covering all groups,
    authorisations, devices, and recent admin changes.
  \item[\textsc{File Events} --- \texttt{splunk\_run\_query}]
    Phase~2C: 7-day file-event timeline including downloads, uploads,
    copies, and shares, with sensitivity labels and file descriptions.
  \item[\textsc{Outbound Email} --- \texttt{splunk\_run\_query}]
    Phase~2B/2C: all external outbound emails sent by the user in the
    past 7~days, with attachment types and external recipients.
  \item[\textsc{Print Events} --- \texttt{splunk\_run\_query}]
    Phase~2D: print activity for the user, covering document names,
    page counts, and printer identifiers.
  \item[\textsc{Sign-in Logs} --- \texttt{entra\_sign\_in\_logs}]
    Phase~2C: authentication and VPN events around the alert to assess
    whether access occurred from an anomalous location or device.
  \item[\textsc{User Activity Dashboard} --- \texttt{splunk\_search\_dashboard}]
    Phase~2C: SIEM activity dashboard to surface cross-event-type
    anomalies.
  \item[\textsc{Endpoint Device Events} --- \texttt{cs\_get\_device\_events}]
    Phase~2D: USB insertions and removable-media events on the user's
    enrolled device (USB whitelist check).
  \item[\textsc{App Events} --- \texttt{splunk\_search\_app\_events}]
    Phase~2D: SharePoint, OneDrive, and collaboration-platform sharing
    and export actions over 7~days.
  \item[\textsc{Domain TI} --- \texttt{ti\_lookup\_domain} (fan-out)]
    Phase~2B: threat-feed lookup per destination domain extracted from
    file and email events (flags personal email services and competitor
    domains).
  \item[\textsc{IP TI} --- \texttt{ti\_lookup\_ip} (fan-out)]
    Phase~2B: threat-feed lookup per external IP from sign-in or query
    results.
  \item[\textsc{External Recipient TI} --- \texttt{ti\_lookup\_email\_address} (fan-out)]
    Phase~2B: threat-feed lookup per external recipient address from
    outbound email, targeting personal or anomalous destination accounts.
\end{description}


\newpage

\end{document}